\documentclass[hidelinks]{article}

    \usepackage[preprint]{neurips_2021}

\usepackage[utf8]{inputenc} 
\usepackage[T1]{fontenc}    
\usepackage{hyperref}       
\usepackage{url}            
\usepackage{booktabs}       
\usepackage{amsfonts}       
\usepackage{nicefrac}       
\usepackage{microtype}      
\usepackage{xcolor}         

\usepackage[inline]{enumitem}
\usepackage{tikz}
\usetikzlibrary{automata,chains}
\usepackage{color}      

\usepackage{xcolor}
\usepackage{multirow}
\usepackage{eucal}
\usepackage{comment}
\usepackage{float}
\usepackage{bm}
\usepackage[fleqn]{amsmath}
\usepackage{subcaption}
\usepackage{mathtools}
\usepackage{xcolor,colortbl}
\usepackage{xspace}

\usepackage{graphicx}
\usepackage{wrapfig}

\def\x{\vecx}
\def\w{\vecw}

\def\Re{\mathbb{R}}

\def\Nat{{\rm I\kern\pIR N}}

\newcommand{\EE}[1]{\exptE\left[#1\right]}

\def\PBE{\overline{\text{PBE}}}
\def\BE{\overline{\text{BE}}}
\def\RVE{\text{R}\overline{\text{VE}}}

\newcommand{\defeq}{\overset{\text{\tiny def}}{=}}

\def\A{{\mathcal{A}}}

\def\E{{\mathcal{E}}}

\def\H{{\mathcal{H}}}

\def\R{{\mathcal{R}}}
\def\S{{\mathcal{S}}}

\newcommand{\States}{\S}
\newcommand{\Actions}{\A}

\def\vec0{{\boldsymbol{0}}}
\def\vecb{{\boldsymbol{b}}}

\def\vecv{{\boldsymbol{v}}}
\def\vecw{{\boldsymbol{w}}}
\def\vecx{{\boldsymbol{x}}}

\def\vecz{{\boldsymbol{z}}}

\def\Xmat{\mathbf{X}}
\def\Dmat{\mathbf{D}}

\newcommand{\Cmat}{\mathbf{C}}
\newcommand{\Amat}{\mathbf{A}}
\newcommand{\inv}{{-1}}
\newcommand{\sneg}{\mathrm{-}}

\newcommand{\beq}{\begin{equation}}
\newcommand{\eeq}{\end{equation}}
\newcommand{\beqa}{\begin{eqnarray}}
\newcommand{\eeqa}{\end{eqnarray}}
\newcommand{\beqan}{\begin{eqnarray*}}
\newcommand{\eeqan}{\end{eqnarray*}}
\newcommand{\ben}{\begin{eqnarray*}}
\newcommand{\een}{\end{eqnarray*}}

\def\tr{^\top\!}

\def\vecw{{\boldsymbol{\bf w}}}
\def\vecz{{\boldsymbol{\bf z}}}
\def\vecx{{\boldsymbol{\bf x}}}

\def\vecv{{\boldsymbol{\bf v}}}
\renewcommand{\EE}[2]{\mathbb{E}_{#1\!\!}\left[#2\right]}

\newcommand{\CEE}[3]{\EE{#1}{{#2}~\middle\vert~{#3}}}
\renewcommand{\CEE}[3]{\EE{#1}{{#2}\mid{#3}}}
\def\CE#1#2{\CEE{\,}{#1}{#2}}
\def\CEb#1#2{\CEE{b}{#1}{#2}}
\def\CEpi#1#2{\CEE{\pi}{#1}{#2}}

\def\E#1{\EE{\,}{#1}}

\def\Eb#1{\EE{b}{#1}}

\def\dtp{\delta_t^\prime}
\def\ztp{\vecz_t^\prime}
\def\dt{\delta_t}
\def\zt{\vecz_t}

\def\ztpp{\vecz_t^\prime}

\newcommand{\vhat}{\hat{v}}
\newcommand{\Glambda}{G^\lambda}

\def\la{($\lambda$)\xspace}
\def\l{$\lambda$\xspace}
\def\a{$\alpha$\xspace}
\def\ze{($\zeta$)\xspace}
\def\lab{($\lambda,\beta$)\xspace}

\title{An Empirical Comparison of Off-policy Prediction Learning Algorithms on the Collision Task}

\author{%
 Sina Ghiassian \\
  University of Alberta\\
  \texttt{ghiassia@ualberta.ca} \\
  \And
  Richard S. Sutton \\
  University of Alberta and DeepMind \\
  \texttt{rsutton@ualberta.ca}
}

\begin{document}

\maketitle

\begin{abstract}
Off-policy prediction---learning the value function for one policy from data generated while following another policy---is one of the most challenging subproblems in reinforcement learning.
This paper presents empirical results with eleven prominent off-policy learning algorithms that use linear function approximation: five Gradient-TD methods, two Emphatic-TD methods, Off-policy TD\la, Vtrace, and versions of Tree Backup and ABQ modified to apply to a prediction setting.
Our experiments used the Collision task, a small idealized off-policy problem analogous to that of an autonomous car trying to predict whether it will collide with an obstacle.
We assessed the performance of the algorithms according to their learning rate, asymptotic error level, and sensitivity to step-size and bootstrapping parameters.
By these measures, the eleven algorithms can be partially ordered on the Collision task.
In the top tier, the two Emphatic-TD algorithms learned the fastest, reached the lowest errors, and were robust to parameter settings.
In the middle tier, the five Gradient-TD algorithms and Off-policy TD\la were more sensitive to the bootstrapping parameter. 
The bottom tier comprised Vtrace, Tree Backup, and ABQ; these algorithms were no faster and had higher asymptotic error than the others.
Our results are definitive for this task, though of course experiments with more tasks are needed before an overall assessment of the algorithms' merits can be made.
\end{abstract}

\section{The Problem of Off-policy Learning}
\label{sct:OffPolicyLearning}

In reinforcement learning, it is not uncommon to learn the value function for one policy while following another policy. 
For example, the Q-learning algorithm (Watkins, 1989; Watkins \& Dayan, 1992) learns the value of the greedy policy while the agent may select its actions according to a different, more exploratory, policy.
The first policy, the one whose value function is being learned, is called the \emph{target policy} while the more exploratory policy generating the data is called the \emph{behavior policy}.
When these two policies are different, as they are in Q-learning, the problem is said to be one of \emph{off-policy learning}, whereas if they are the same, the problem is said to be one of \emph{on-policy learning}. The former is `off' in the sense that the data is from a different source than the target policy, whereas the latter is from data that is `on' the policy.
Off-policy learning is more difficult than on-policy learning and subsumes it as a special case.

One reason for interest in off-policy learning is that it provides a clear way of intermixing exploration and exploitation.
The classic dilemma is that an agent should always \emph{exploit} what it has learned so far---it should take the best actions according to what it has learned---but it should also always \emph{explore} to find actions that might be superior.
No agent can simultaneously behave in both ways.
However, an off-policy algorithm like Q-learning can, in a sense, pursue both goals at the same time. The behavior policy can explore freely while the target policy can converge to the fully exploitative, optimal policy independent of the behavior policy's explorations.

Another appealing aspect of off-policy learning is that it enables learning about many policies in parallel. Once the target policy is freed from behavior, there is no reason to have a single target policy. With off-policy learning, an agent could simultaneously learn how to optimally perform many different tasks (as suggested by Jaderberg et al.\ (2016) and Rafiee et al.\ (2019)). Parallel off-policy learning of value functions has even been proposed as a way of learning general, policy-dependent, world knowledge (e.g., Sutton et al., 2011; White, 2015; Ring, in prep).

Finally, note that numerous ideas in the machine learning literature rely on effective off-policy learning, including the learning of temporally-abstract world models (Sutton, Precup, \& Singh, 1999), predictive representations of state (Littman, Sutton, \& Singh, 2002; Tanner \& Sutton, 2005), auxiliary tasks (Jaderberg et al., 2016), life-long learning (White, 2015), and learning from historical data (Thomas, 2015).

Many off-policy learning algorithms have been explored in the history of reinforcement learning. Q-learning (Watkins, 1989; Watkins \& Dayan, 1992) is perhaps the oldest.
In the 1990s it was realized that combining off-policy learning, function approximation, and temporal-difference (TD) learning risked instability (Baird, 1995).
Precup, Sutton, and Singh (2000) introduced off-policy algorithms with importance sampling and eligibility traces, as well as tree backup algorithms, but did not provide a practical solution to the risk of instability. Gradient-TD methods (see Maei, 2011; Sutton et al., 2009) assured stability by following the gradient of an objective function, as suggested by Baird (1999). Emphatic-TD methods (Sutton, Mahmood, \& White, 2016) reweighted updates in such a way as to regain the convergence assurances of the original on-policy TD algorithms. These methods had convergence guarantees, but no assurances that they would be efficient in practice. Other off-policy algorithms, including Retrace (Munos et al., 2016), Vtrace (Espeholt et al., 2018), and ABQ (Mahmood, Yu, \& Sutton, 2017) were developed recently to overcome difficulties encountered in practice.

As more off-policy learning methods were developed, there was a need to compare them systematically.
The earliest systematic study was that by Geist and Scherrer (2014). Their experiments were on random MDPs and compared eight major off-policy algorithms.
A few months later, Dann, Neumann, and Peters (2014) published a more in-depth study with one additional algorithm (an early Gradient-TD algorithm) and six test problems including random MDPs. 
Both studies considered off-policy problems in which the target and behavior policies were  given and stationary. Such \emph{prediction} problems allow for relatively simple experiments and are still challenging (e.g., they involve the same risk of instability).
Both studies also used linear function approximation with a given feature representation.
The algorithms studied by Geist and Scherrer (2014), and by Dann, Neumann, and Peters (2014) can be divided into those whose per-step complexity is linear in the number of parameters, like TD\la, and methods whose complexity is quadratic in the number of parameters (proportional to the square of the number of parameters), like Least Squares TD\la (Bradtke \& Barto, 1996; Boyan, 1999). Quadratic-complexity methods avoid the risk of instability, but cannot be used in learning systems with large numbers (e.g., millions) of weights.
A third systematic study, by White and White (2016), excluded quadratic-complexity algorithms, but added four additional linear-complexity algorithms.

The current paper is similar to previous studies in that it treats prediction with linear function approximation, and similar to the study by White and White (2016) in restricting attention to linear complexity algorithms. Our study differs from earlier studies in that it treats more algorithms and does a deeper empirical analysis on a single problem, the Collision task.  
The additional algorithms are the prediction variants of Tree Backup\la (Precup, Sutton, \& Singh, 2000), Retrace\la (Munos et al., 2016), ABQ($\zeta$) (Mahmood, Yu, \& Sutton, 2017), and TDRC\la (Ghiassian et al., 2020).
Our empirical analysis is deeper primarily in that we examine and report the dependency of all eleven algorithms' performance on all of their parameters. 
This level of detail is needed to expose our main result, an overall ordering of the performance of off-policy algorithms on the Collision task.
Our results, though limited to this task, are a significant addition to what is known about the comparative performance of off-policy learning algorithms.


\section{Formal Framework}
\label{sct:TheProblemOfOffpolicyPrediction}

In this section, we formally explain the framework of off-policy prediction learning with linear function approximation.
An agent and environment interact at discrete time steps, $t=0, 1, 2, \ldots$.
The environment is a Markov Decision Process (MDP) with state $S_t\in\S$ at time step $t$.
At each time step, the agent chooses an action $A_t\in\A$ with probability $b(a|s)$, where the function $b:~\A \times \S \rightarrow [0, 1]$ with $\sum_{a\in\A} b(a|s) =1, \forall s\in\S$,  is called the \emph{behavior} policy because it determines the agent's behavior. 
After taking action $A_t$ in state $S_t$, the agent receives from the environment a numerical reward $R_{t+1}\in\R\subset\Re$ and the next state $S_{t+1}$. In general the reward and next state are stochastically jointly determined by the current state and action.

In prediction learning, we estimate for each state the expected discounted sum of future rewards, given that actions are taken according to a different policy $\pi$, called the \emph{target} policy (because learning its values is the target of our learning).
For simplicity, both target and behavior policies are assumed here to be known and static, although of course in many applications of interest one or the other may be changing.
The discounted sum of future rewards at time $t$ is called the \emph{return} and denoted $G_t$:
\begin{align}
G_t \defeq R_{t+1} + \gamma R_{t+2} + \gamma^2 R_{t+3} + \cdots \nonumber
\end{align}
The expected return when starting from a state and following a specific policy thereafter is called the \emph{value} of the state under the policy.
The \emph{value function} $v_\pi:\S\rightarrow\Re$ for a policy $\pi$ takes a state as input and returns the value of that state: 
 \begin{align}
 v_\pi(s) \defeq \CE{G_t}{S_t\!=\!s, A_{t:\infty}\!\sim\pi}.
 \end{align}
 
Prediction learning algorithms seek to learn an estimate $\vhat: \S\rightarrow\Re$ that approximates the true value function $v_\pi$.
In many problems $\S$ is large and an exact approximation is not possible even in the limit of infinite time and data.
Many parametric forms are possible, including deep artificial neural networks, but of particular interest, and our exclusive focus here, is the linear form: 
\beq
\vhat(s, \vecw)\defeq \vecw\tr\vecx(s), 
\eeq
where $\vecw \in \Re^d$ is a learned weight vector and $\vecx(s)\in \Re^d, \forall s\in\S$ is a set of given feature vectors, one per state, where $d \ll |\S|$.


\section{Algorithms}
\label{sct:TemporalDifferenceAlgorithmsForOffPolicyPrediction}

In this section, we briefly introduce the eleven algorithms used in our empirical study.
These eleven are intended to include all the best candidate algorithms for off-policy prediction learning with linear function approximation.
The complete update rules of all algorithms and additional technical discussion can be found in Appendix~\ref{app:PseudoCodes} and Appendix~\ref{app:TemporalDifferenceAlgorithmsForOffPolicyPrediction} respectively.

\emph{Off-policy TD\la} (Precup, Sutton, \& Dasgupta, 2001) is the off-policy variant of the original TD\la algorithm (Sutton, 1988) that uses importance sampling to reweight the returns and account for the differences between the behavior and target policies. This algorithm has just one set of weights and one step size parameter.

Our study includes five algorithms from the Gradient-TD family.
\emph{GTD\la} and \emph{GTD2\la} are based on algorithmic ideas introduced by Sutton et al.,\ (2009), then extended to eligibility traces by Maei (2011).
\emph{Proximal GTD2\la} (Mahadevan et al., 2014; Liu et al., 2015; Liu et al., 2016) is a ``mirror descent'' version of GTD2 using a saddle-point objective function. 
These algorithms approximate stochastic gradient descent (SGD) on an alternative objective function, the mean squared projected Bellman error.
\emph{HTD\la} (Hackman, 2012; White \& White, 2016) is a ``hybrid'' of GTD\la and TD\la which becomes equivalent to classic TD\la where the behavior policy coincides with the target policy.
\emph{TDRC\la} is a promising recent variant of GTD\la that adds regularization.
All these methods involve an additional set of learned weights (beyond that used in $\vhat$) and a second step-size parameter, which can complicate their use in practice. 
TDRC\la offers a standard way of setting the second step-size parameter, which makes this less of an issue.
All of these methods are guaranteed to converge with an appropriate setting of their two step-size parameters.

Our study includes two algorithms from the Emphatic-TD family.
Emphatic-TD algorithms attain stability by up- or down-weighting the updates made on each time step by Off-policy TD\la. If this variation in the emphasis of updates is done in just the right way, stability can be guaranteed with a single set of weights and a single step-size parameter. The original emphatic algorithm, \emph{Emphatic TD\la}, was introduced by Sutton, Mahmood, and White (2016). The variant \emph{Emphatic TD($\lambda,\beta$)}, introduced by Hallak et al.,\ (2016), has an additional parameter, $\beta\in[0, 1]$, intended to reduce variance.

The final three algorithms in our study can be viewed as different attempts to address the problem of large variations in the product of importance sampling ratios. 
If this product might become large, then the step-size parameter must be set small to ensure there is no overshoot---and then learning may be slow. All these methods attempt to control the importance sampling product by changing the bootstrapping parameter from step to step (Yu, Mahmood, \& Sutton, 2018).
Munos et al.,\ (2016) proposed simply putting a cap on the importance sampling ratio at each time step; they explored the theory and practical consequences of this modification in a control context with their Retrace algorithm.
\emph{Vtrace\la} (Espeholt et al., 2018) is a modification of Retrace to make it suitable for prediction rather than control. 
Mahmood et al.,\ (2017) developed a more flexible algorithm that achieves a similar effect.
Their algorithm was also developed for control; to apply the idea to prediction learning we had to develop a nominally new algorithm, \emph{ABTD\ze}, that naturally extends ABQ\ze from control to prediction. A full development of ABTD\ze can be found in Appendix~\ref{app:DerivationForVariableLambda}.
Finally, \emph{Tree Backup\la} (Precup, Sutton, \& Singh, 2000) reduces the effective \l by the probability of the action taken on each time step.
Each of these algorithms (or their control predecessors) have been shown to be very effective on specific problems.


\section{Collision Task}
\label{sct:SpecifyingTheCollisionTask}

\def\forward{\texttt{forward}\xspace}
\def\turnaway{\texttt{turnaway}\xspace}

\begin{figure}[hb]
      \centering
      \includegraphics[width=0.8\linewidth]{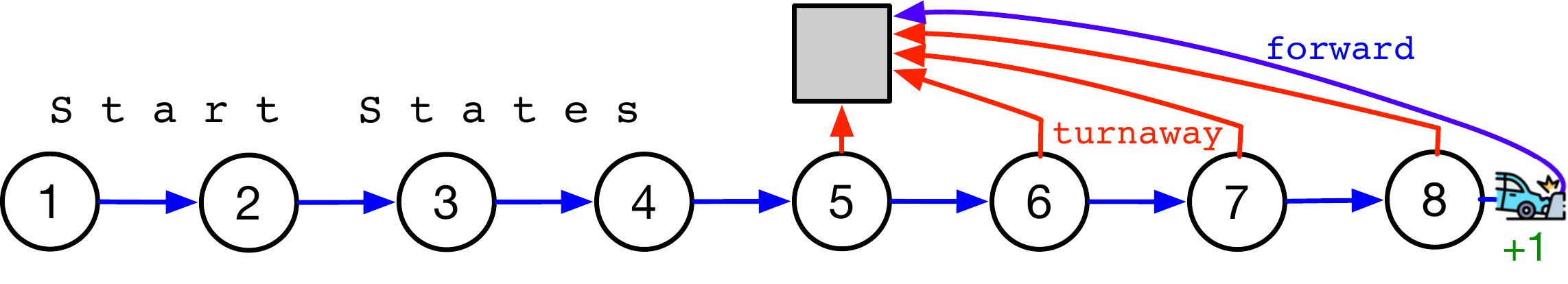}
      \caption{The Collision task. Episodes start in one of the first four states and end when the \forward action is taken from the eighth state, causing a crash and a reward of 1, or when the \turnaway action is taken in one of the last four states.}
      \label{fig:-2_4-CollisionTask-Fig-v3}
\end{figure}

The Collision task is an idealized off-policy prediction-learning task. 
A vehicle moves along an eight-state track towards an obstacle with which it will collide if it keeps moving forward. In this episodic task, each episode begins with the vehicle in one of the first four states (selected at random with equal probability). In these four states, \forward is the only possible action whereas, in the last four states, two actions are possible: \forward and \turnaway (see Figure \ref{fig:-2_4-CollisionTask-Fig-v3}). The \forward action always moves the vehicle one state further along the track; if it is taken in the last state, then a collision is said to occur, the reward is 1, and the episode ends. The \turnaway action causes the vehicle to ``turn away'' from the wall, which also ends the episode, except with a reward of zero. The reward is also zero on all earlier, non-terminating transitions. In an episodic task like this the return is accumulated only up to the end of the episode. After termination, the next state is the first state of the next episode, selected randomly from the first four as specified above.

\begin{wrapfigure}{r}{0.35\textwidth}
  \begin{center}
    \includegraphics[width=0.35\textwidth]{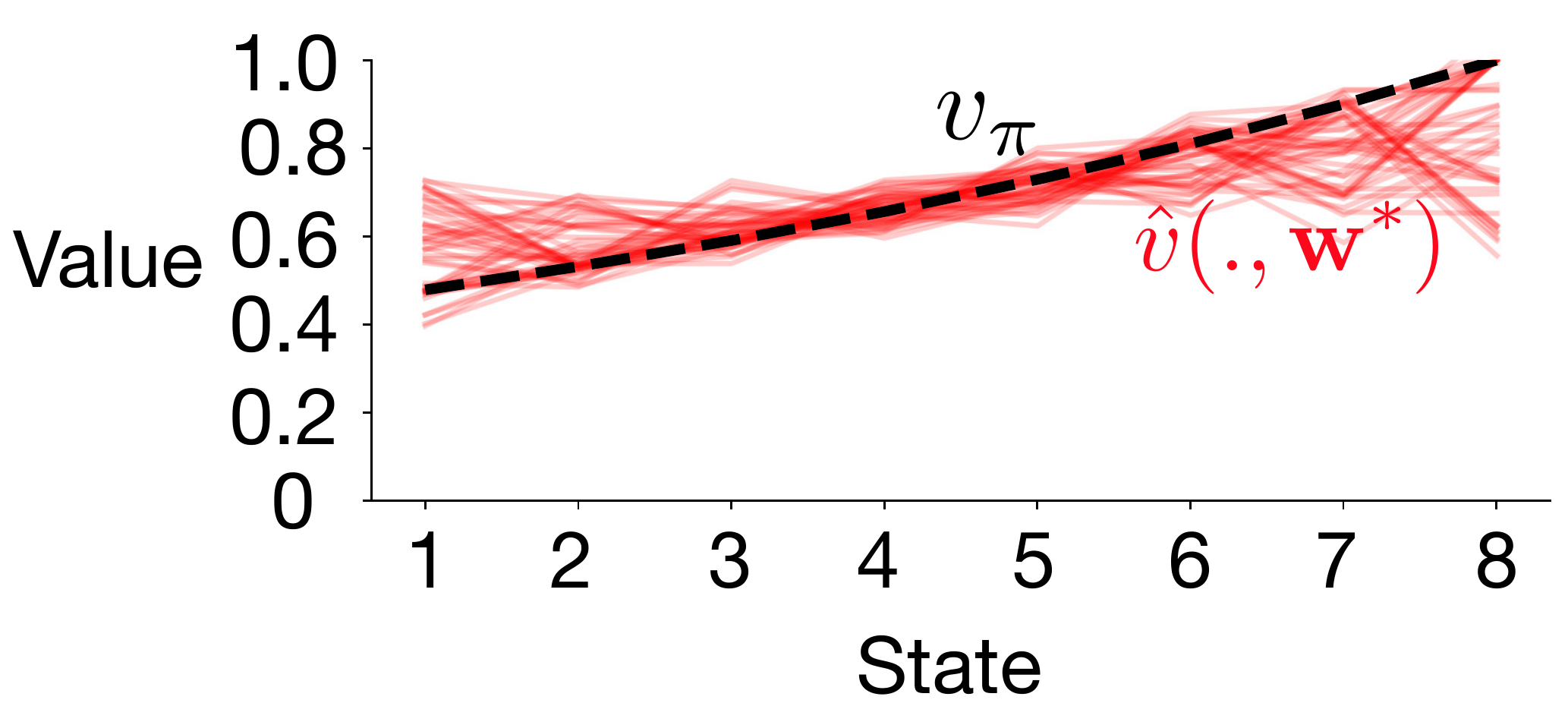}
  \end{center}
  \caption{The ideal value function, $v_\pi$, and the best approximate value functions, $\vhat$, for 50 different feature representations.}
      \label{fig:-1_5-Collision-value_functions}
\end{wrapfigure}

The target policy on this task is to always take the forward action, $\pi(\forward|s)=1, \forall s\in\S$, whereas the behavior policy is to take the two actions (where available) with equal probability, $b(\forward|s)=b(\turnaway|s)=0.5, \forall s\in\{5, 6, 7, 8\}$. The problem is discounted with a discount rate of $\gamma=0.9$. As always, we are seeking to learn the value function for the target policy, which in this case is $v_\pi(s)=\gamma^{8-s}$. This function is shown as a dotted black line in Figure \ref{fig:-1_5-Collision-value_functions}. The thin red lines show approximate value functions $\vhat\approx v_\pi$, using various feature representations, as we discuss shortly below.

This idealized task is roughly analogous to and involves some similar issues as real-world autonomous driving problems, such as exiting a parallel parking spot without hitting the car in front of you, or learning how close you can get to other cars without risking collisions. In particular, if these problems can be treated as off-policy learning problems, then solutions can potentially be learned with fewer collisions. In this paper, we are testing the efficiency of various off-policy prediction-learning algorithms at maximizing how much they learn from the same number of collisions.

Similar problems have been studied using mobile robots.
For example, White (2015) used off-policy learning algorithms running on an iRobot Create to predict collisions as signaled by activation of the robot's front bumper sensor.
Rafiee et al. (2019) used a Kobuki robot to not only anticipate collisions, but to turn away from anticipated collisions before they occurred.
Modayil and Sutton (2014) trained a custom robot to predict motor stalls and turn off the motor when a stall was predicted.

We artificially introduce function approximation into the Collision task. Although a tabular approach is entirely feasible on this small problem, it would not be on the large problems of interest. In real applications, the agent would have sensor readings, or more generally a state representation, which would help it distinguish between states.
We simulate such sensor readings in the Collision task by randomly assigning to each of the eight states a binary feature vector $\x(s)\in\{0,1\}^d, \forall s\in\{1..8\}$. 
We chose $d=6$, so that is was not possible for all eight of the feature vectors (one per state) to be linearly independent. In particular, we chose all eight feature vectors to have exactly three 1s and three 0s, with the location of the 1s for each state being chosen randomly.

\def\MSVEm{\overline{\text{VE}}}
\def\MSVE{$\overline{\text{VE}}$\xspace}
Because the feature vectors are linearly dependent, it is not possible in general for a linear approximation, $\vhat(s,\w)=\w\tr\x$, to equal to $v_\pi(s)$ at all eight states of the Collision task.
This, in fact, is the sole reason the red approximate value functions in Figure \ref{fig:-1_5-Collision-value_functions} do not exactly match $v_\pi$. Given a feature representation $\x:\S\rightarrow\Re^d$, a linear approximate value function is completely determined by its weight vector $\w\in\Re^d$. The quality of that approximation is assessed by its squared error at each state, weighted by how often each state occurs:
\beq
\MSVEm(\w)=\sum_{s\in\S}\mu_b(s)\bigl[\hat{v}(s,\w) - v_\pi(s)  \bigr]^2, 
\eeq
where $\mu_b(s)$ is the state distribution, the fraction of time steps in which $S_t=s$, under the behavior policy (here $\mu_b$ was approximated from visitation counts from one million sample time steps). The value functions shown by red lines in Figure \ref{fig:-1_5-Collision-value_functions} are for $\w^*$, the weight vector that minimizes $\MSVEm(\w)$, with each line corresponding to a different randomly selected feature representation as described earlier. For these value functions, $\MSVEm(\w^*)\approx 0.05$.
All the code for the Collision task and these experiments is available at: \href{https://github.com/sinaghiassian/OffpolicyAlgorithms}{https://github.com/sinaghiassian/OffpolicyAlgorithms}.


\section{Experiment}
\label{sct:The Experiment}

The Collision task, in conjunction with its behavior policy, was used to generate 20,000 time steps, comprising one \emph{run}, and then this was repeated for a total of 50 independent runs.
Each run also used a different feature representation randomly generated as described in the previous section.
The eleven learning algorithms were then applied to the 50 runs, each with a range of parameter values; each combination of algorithm and parameter settings is termed an \emph{algorithm instance}.
A list of all parameter settings used can be found in Table~\ref{tab:params-tbl} of Appendix~\ref{app:ParameterSettings}.
They included 12 values of \l, 19 values of \a, 15 values of $\eta$ (for the Gradient-TD family), six values of $\beta$ (for ETD($\lambda, \beta$)), and 19 values of $\zeta$ (for ABTD\ze), for approximately 20,000 algorithm instances in total.
\begin{wrapfigure}{r}{0.4\textwidth}
  \begin{center}
    \includegraphics[width=0.4\textwidth]{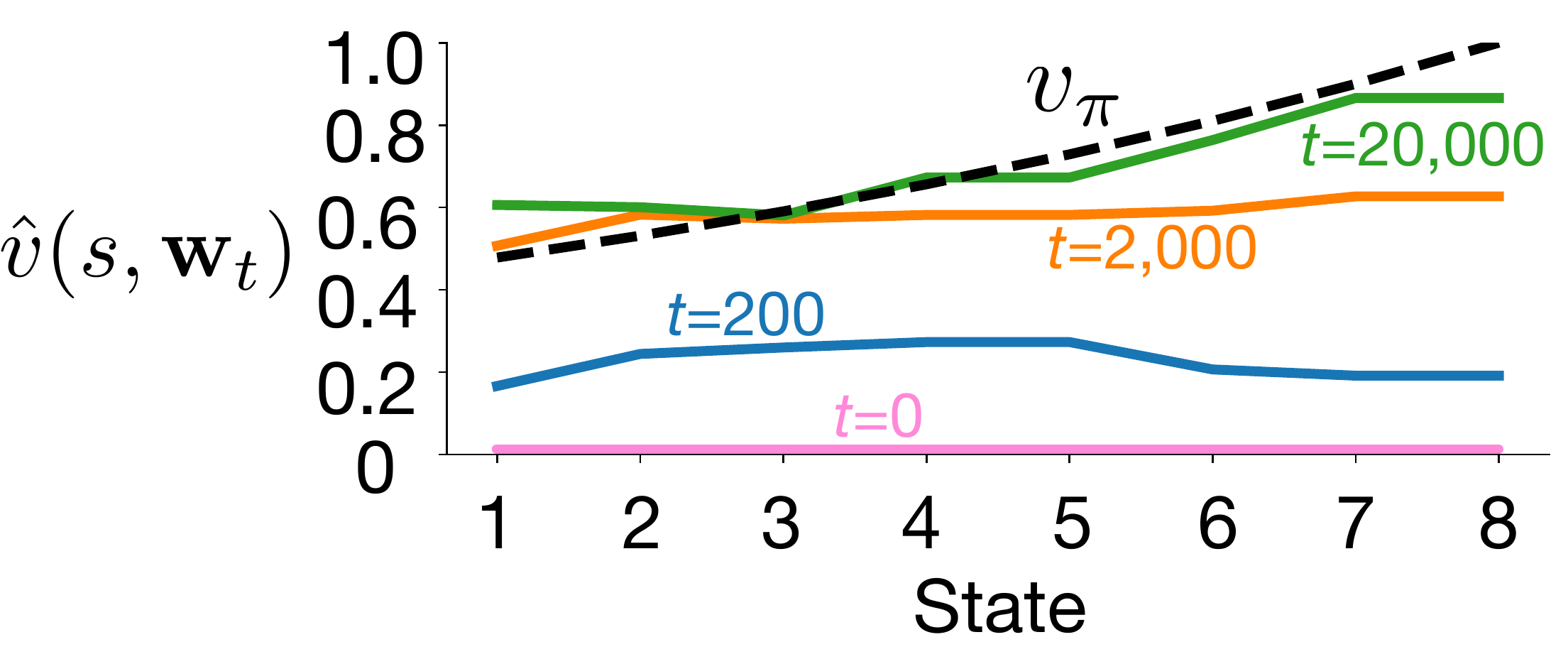}
  \end{center}
  \caption{An example of the approximate value function, $\vhat$, being learned over time.}
      \label{fig:-1-Collision-value_functions}
\end{wrapfigure}
In each run, the weight vector was initialized to $\w_0=\vec0$ and then updated at each step by the algorithm instance to produce a sequence of $\w_t$. At each step we also computed and recorded $\MSVEm(\w_t)$.

With a successful learning procedure, we expect the value function to evolve over time as in Figure~\ref{fig:-1-Collision-value_functions}.
The approximate value function starts at $\vhat(s,\vec0)=0$, as shown by the pink line, then moves toward positive values, as shown by the blue and orange lines.
Finally, the learned value function slants and comes to closely approximate the true value function, though always with some residual error due to the limited feature representation, as shown by the green line (and also by all the red lines in Figure \ref{fig:-1_5-Collision-value_functions}).

\newpage

\begin{wrapfigure}{r}{0.5\textwidth}
  \begin{center}
    \includegraphics[width=0.5\textwidth]{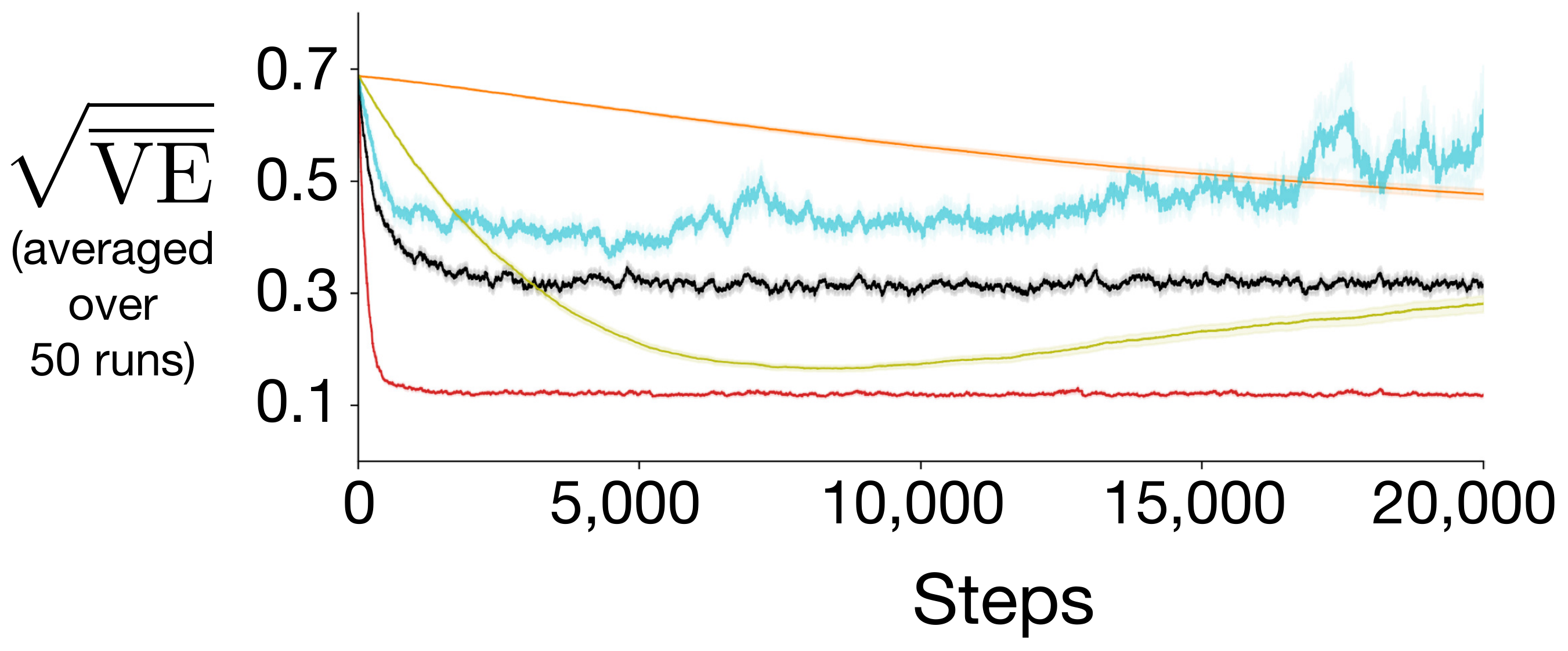}
  \end{center}
  \caption{Learning curves illustrating the range of things that can happen during a run. The average error over the 20,000 steps is a good combined measure of learning rate and asymptotic error.
      }
      \label{fig:0-Collision-sample_learning_curves}
\end{wrapfigure}
Figure~\ref{fig:0-Collision-sample_learning_curves} shows learning curves illustrating the range of things that happened in the experiment.
Normally, we expect \MSVE to decrease over the course of the experiment, starting at $\MSVEm(\vec0)\approx 0.7$ and falling to some minimum value, as in the red and black lines in Figure~\ref{fig:0-Collision-sample_learning_curves} (these and all other data are averaged over the fifty runs).
If the primary step-size parameter, \a, is small, then the learning may be slow and incomplete by the end of the runs, as in the orange line. A larger step-size parameter may be faster, but, if it is too large, then divergence can occur, as in the blue line. For one algorithm, Proximal GTD2\la, we found that the error dipped low and then leveled off at a higher level, as in the olive line. 


\section{Main Results: A Partial Order over Algorithms}
\label{sct:Results}

\begin{figure}[t]
      \includegraphics[width=\textwidth]{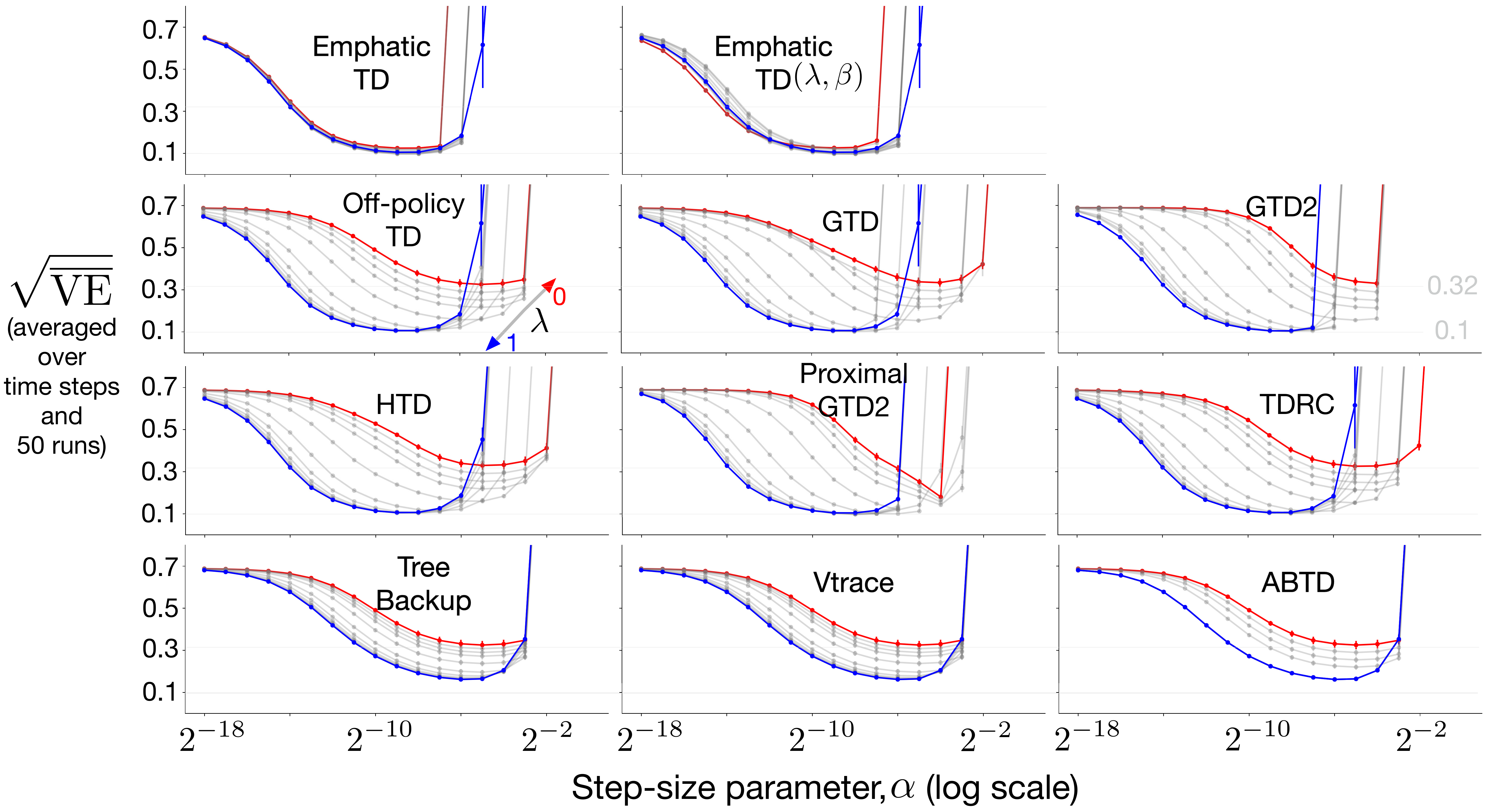}
      \caption{
      Main results: Performance of all algorithms on the Collision task as a function of their parameters \a and \l. The top tier algorithms (top row) attained a low error ($\approx$0.1) at all \l values. The middle tier of six algorithms attained a low error for $\lambda=1$, but not for $\lambda=0$. And the bottom-tier of three algorithms were unable to reach an error of $\approx$0.1 at any \l value.}
      \label{fig:5_4-Collision-Lambda-Sensitivity-Curves}
\end{figure}

As an overall measure of the performance of an algorithm instance, we take its learning curve over 50 runs, as in Figure~\ref{fig:0-Collision-sample_learning_curves}, and then average it across the 20,000 steps. In this way, we reduce all the data for an algorithm instance to a single number that summarizes its performance. These numbers appear as points in our main results figure, Figure \ref{fig:5_4-Collision-Lambda-Sensitivity-Curves}. Each panel of the figure is devoted to a single algorithm. 

For example, performance numbers for instances of Off-policy TD\la are shown as points in the left panel of the second row of Figure \ref{fig:5_4-Collision-Lambda-Sensitivity-Curves}. This algorithm has two parameters, the step-size parameter, \a, and the bootstrapping parameter, \l. The points are plotted as a function of \a, and  points with the same \l value are connected by lines. The blue line shows the performances of the instances of Off-policy TD\la with $\lambda=1$, the red line shows the performances with $\lambda=0$, and the gray lines show the performances with intermediate $\lambda$s. Note that all the lines are U-shaped functions of \a, as is to be expected; at small \a learning is too slow to make much progress, and at large \a there is overshoot and divergence, as in the blue line in Figure \ref{fig:0-Collision-sample_learning_curves}. For each point, the standard error over the 50 runs is also given as an error bar, though these are too small to be seen in all except the rightmost points of each line where the step size was highest and divergence was common. Except for these rightmost points, almost all visible differences are statistically significant.

First focus on the blue line (of the left panel on the second row of Figure \ref{fig:5_4-Collision-Lambda-Sensitivity-Curves}), representing the performances of Off-policy TD\la with $\lambda=1$. There is a wide sweet spot, that is, there are many intermediate values of \a at which good performance (low average error) is achieved. Note that the step-size parameter \a is varied over a wide range, with logarithmic steps. The minimal error level of about 0.1 was achieved over four or five powers of two for \a. This is the primary measure of good performance that we look for in these data: low error over a wide range of parameter values.

Now contrast the blue line with the red and gray lines (for Off-policy TD\la in the left panel of the second row of Figure \ref{fig:5_4-Collision-Lambda-Sensitivity-Curves}). Recall that the blue line is for $\lambda=1$, the red line is for $\lambda=0$, and the gray lines are for intermediate values of \l. First note that the red line shows generally worse performance; the error level at $\lambda=0$ was higher, and its range of good \a values was slightly smaller (on a logarithmic scale). The intermediate values of \l all had performances that were between the two extremes. Second, the sweet spot (the best \a value) consistently shifted right, toward higher \a, as \l was decreased from 1 toward 0.

Now, armed with a thorough understanding of the Off-policy TD\la panel, consider the other panels of Figure \ref{fig:5_4-Collision-Lambda-Sensitivity-Curves}.
Overall, there are a lot of similarities between the algorithms and how their performances varied with \a and \l. For all algorithms, error was lower for $\lambda=1$ (the blue line) than for $\lambda=0$ (the red line). Bootstrapping apparently confers no advantage in the Collision task for any algorithm. 

The most obvious difference between algorithms is that the performance of the two Emphatic-TD algorithms varied relatively little as a function of \l; their blue and red lines are almost on top of one another, whereas those of all the other algorithms are qualitatively different.
Moreover, the emphatic algorithms generally performed as well as or better than the other algorithms. At $\lambda=1$, the emphatic algorithms reached the minimal error level of all algorithms ($\approx$0.1), and their ranges of good \a values was just as wide as that of the other algorithms. While at $\lambda=0$, the best errors of the emphatic algorithms were qualitatively better than those of the other algorithms. The minimal $\lambda=0$ error level of the emphatic algorithms was about 0.15, as compared to approximately 0.32 (shown as a second thin gray line) for all the other algorithms (except Proximal GTD2, a special case that we consider later). Moreover, for the emphatic algorithms the sweet spot for \a shifted little as \l varied. The shift was markedly less than for the six algorithms in the middle two rows of Figure \ref{fig:5_4-Collision-Lambda-Sensitivity-Curves}. The lack of an interaction between the two parameter values is another potential advantage of the emphatic algorithms.

The lowest error level for eight of the algorithms was $\approx$0.1 (shown as a thin gray line), and for the other three algorithms the best error was higher, $\approx$0.16. The differences between the eight and the three were highly statistically significant, whereas the differences within the two groups were negligible. The three algorithms that performed worse than the others were Tree Backup\la, Vtrace\la, and ABTD\ze---shown in the bottom row of Figure \ref{fig:5_4-Collision-Lambda-Sensitivity-Curves}. The difference was only for large $\lambda$s; at $\lambda=0$ these three algorithms reached the same error level ($\approx$0.32) as the other non-emphatic algorithms. The three worse algorithms' range of good \a values was also slightly smaller than for the other algorithms (with the partial exception, again, of Proximal GTD2\la). A mild strength of the three is that the best \a value shifted less as a function of \l than for the other six non-emphatic algorithms. Generally, the performances of these three algorithms in Figure \ref{fig:5_4-Collision-Lambda-Sensitivity-Curves} look very similar as a function of parameters. An interesting difference is that for ABTD\ze, we only see three gray curves, whereas for the other two algorithms we see seven. For ABTD\ze there is no \l parameter, but the parameter $\zeta$ plays the same role. In our experiment, ABTD\ze performed identically for all $\zeta$ values greater than 0.5; four gray lines with different $\zeta$ values are hidden behind ABTD's blue curve.

In summary, our main result is that on the Collision task the performances of the eleven algorithms fell into three groups, or tiers.
In the top tier are the two Emphatic-TD algorithms, which performed well and almost identically at all values of \l and significantly better than the other algorithms at low \l. Although this difference did not affect best performance here (where $\lambda=1$ is best), the ability to perform well with bootstrapping is expected to be important on other tasks.
In the middle tier are Off-policy TD\la and all the Gradient-TD algorithms including HTD\la, all of which performed well at $\lambda=1$ but less well at $\lambda=0$.
Finally, in the bottom tier are Tree Backup\la, Vtrace\la, and ABTD\la, which performed very similarly and not as well as the other algorithms at their best parameter values. 
All of these differences are statistically significant, albeit specific to this one task.
In Figure \ref{fig:5_4-Collision-Lambda-Sensitivity-Curves} the three tiers are the top row, the two middle rows, and the bottom row.

In the next two sections we take a closer look at two of the tiers to find differences within them.


\section{Emphatic TD\la vs. Emphatic TD\lab}
\label{sct:EmphaticTDVsEmphaticTDBeta}

In this section, the effect of the $\beta$ parameter of Emphatic TD\lab on the algorithm's performance in the full bootstrapping case is analyzed.
We focus on the full bootstrapping case ($\lambda=0$) because this is where the largest differences were observed in the previous section.
The results in Figure~\ref{fig:5_4-Collision-Lambda-Sensitivity-Curves} for Emphatic TD\lab are with its best values of $\beta$.
The curves shown in Figure~\ref{fig:5_4-Collision-Lambda-Sensitivity-Curves}, are for the best values of $\beta$; meaning that, for each \l, we found the combination of \a and $\beta$ that resulted in the minimum average error, fixed $\beta$, and plotted the sensitivity for that fixed $\beta$ over the step-size parameter.
Here, we show how varying $\beta$ affects performance.

\begin{wrapfigure}{r}{0.5\textwidth}
  \begin{center}
    \includegraphics[width=0.5\textwidth]{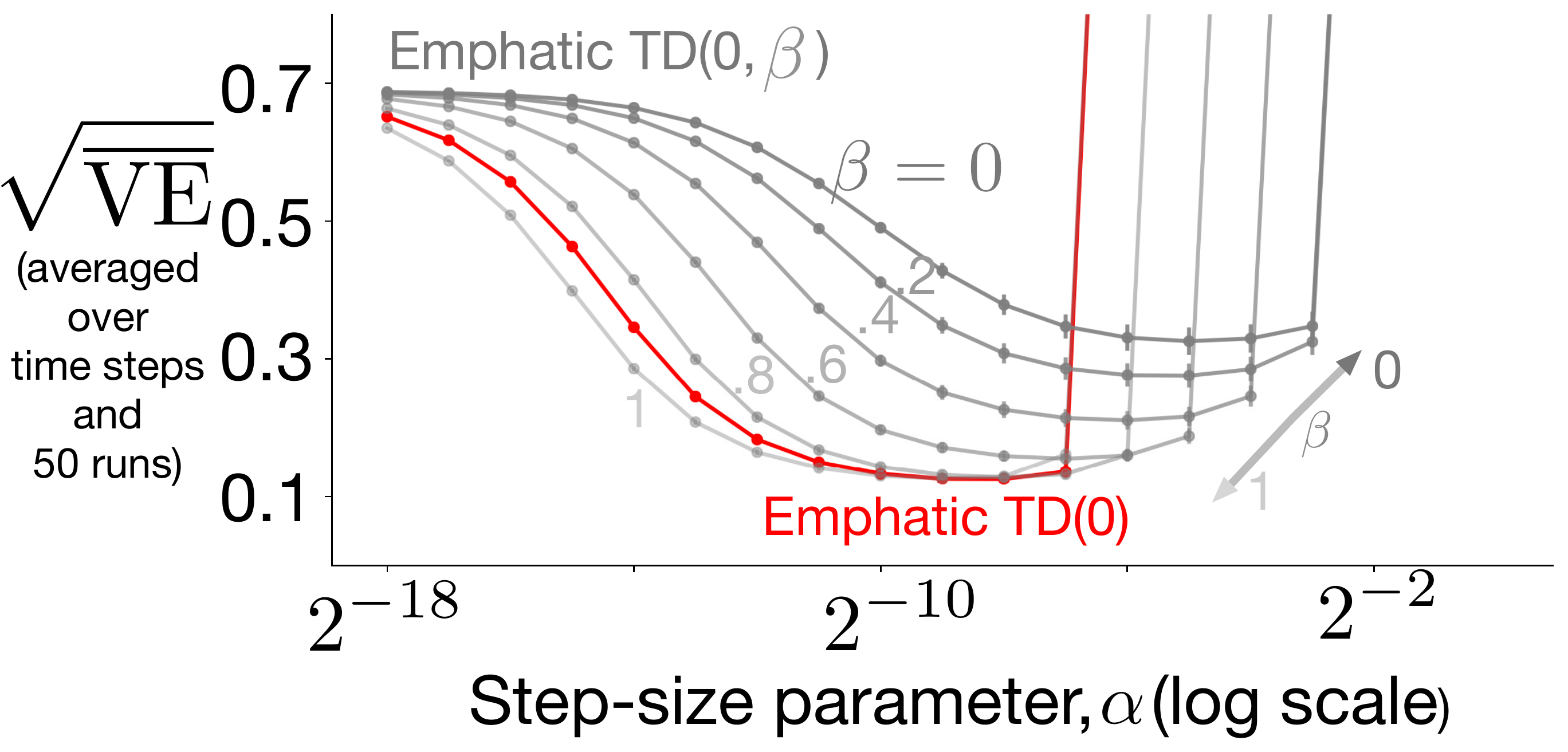}
  \end{center}
  \caption{Detail on the performance of Emphatic TD\lab at $\lambda=0$. Note that Emphatic TD\la is equivalent to Emphatic TD($\lambda, \gamma$), and here $\gamma=0.9$. The flexibility provided by $\beta$ does not help on the Collision task.
      }
      \label{fig:2-Collision-Emphatics}
\end{wrapfigure}
The error of Emphatic TD(0), and Emphatic TD(0,$\beta$) for various values of $\alpha$ and $\beta$ are shown in  Figure~\ref{fig:2-Collision-Emphatics}.
We see that both algorithms performed similarly well on the Collision task, meaning that they both had a wide sensitivity curve and reached the same ($\approx$0.1) error level.
Notice that, as $\beta$ increased, the sensitivity curve for Emphatic TD(0,$\beta$) shifted to left and the bottom of the sensitivity curve shifted down.
With $\beta=0$, Emphatic TD\lab, reduces to TD\la.
With $\beta=0.8$, and $\beta=1$, Emphatic TD\lab reached the same error level as Emphatic TD\la.
With $\beta=\gamma$, Emphatic TD\lab reduces to Emphatic TD\la.
This explains why the red curve is between the $\beta=0.8$ and $\beta=1$ curves.

The results make it clear that the superior performance of emphatic methods are almost entirely due to the basic idea of emphasis; the additional flexibility provided by $\beta$ of the Emphatic TD($\lambda,\beta$) was not important on the Collision problem.


\section{Assessment of Gradient-TD Algorithms}
\label{sct:AssessmentOfGradietTDAlgorithms}

\begin{figure}[t]
      \centering
      \includegraphics[width=0.70\linewidth]{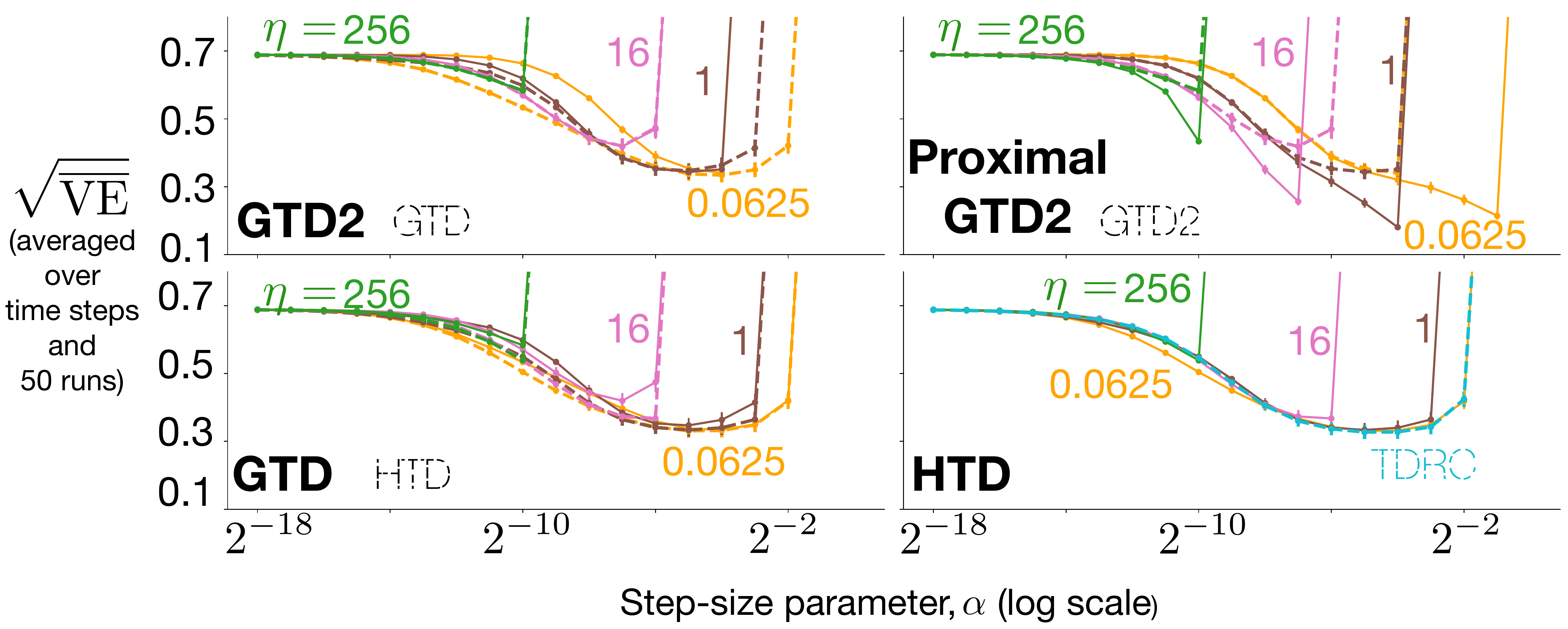}
      \caption{Detail on the performance of Gradient-TD algorithms at $\lambda=0$. Each algorithm has a second step-size parameter, scaled by $\eta$. A second algorithm's performance is also shown in each panel, with dashed lines, for comparison.}
      \label{fig:3-Collision-Gradients}
\end{figure}

We study how the $\eta$ parameter of Gradient-TD algorithms affects performance in the case of full bootstrapping (the second step size, $\alpha_\vecv$, is equal to $\eta \times \alpha$).
Previously, in Figure~\ref{fig:5_4-Collision-Lambda-Sensitivity-Curves} we looked at the results with the best values of $\eta$ for each \l; meaning that for each \l, first the combination of $\alpha$ and $\eta$ that resulted in the lowest average \MSVE was found and then sensitivity to step size was plotted for that specific value of $\eta$ that minimized average \MSVE.
Sensitivity to step size for various values of $\eta$ for $\lambda=0$ are shown in Figure~\ref{fig:3-Collision-Gradients}.
First focus on the upper left panel.
Each panel shows the result of two Gradient-TD algorithms for various $\eta$.
One main algorithm, shown with solid lines, and another additional algorithm shown with dashed lines for comparison.
The upper left panel shows the parameter sensitivity for GTD2(0), for four values of $\eta$, and additionally it shows GTD(0) results as dashed lines for comparison (for results with more values of $\eta$ see Appendix~\ref{app:AdditionalResults}).
The color for each value of $\eta$ is consistent within and across the four panels, meaning that for example, $\eta=256$ is shown in green in all panels, either as dashed or solid lines.
For all parameter combinations, GTD errors were lower than (or similar to) GTD2 errors.
With two smaller values of $\eta$ (1 and 0.0625) GTD had a wider and lower sensitivity curve than GTD2, which means GTD was easier to use than GTD2.

Let us now move on to the upper right panel of Figure~\ref{fig:3-Collision-Gradients}.
Proximal GTD2 had the most distinctive behavior among  Gradient-TD algorithms.
As we previously observed in Figure~\ref{fig:0-Collision-sample_learning_curves}, it is the only algorithm that in some cases had a ``bounce''; its error dipped down at first and then moved back up.
With $\lambda=0$, it sometimes converged to an error that was lower than all other Gradient-TD algorithms.
Proximal GTD2 was more sensitive to the choice of step size than all Gradient-TD algorithms except GTD2.
Proximal GTD2 had a lower error and a wider sensitivity curve than GTD2. To see this, compare the dotted and solid lines in the upper right panel of Figure~\ref{fig:3-Collision-Gradients}.

Moving on to the lower left panel, we see that GTD and HTD performed similarly.
Their sensitivity curves were similarly wide but HTD reached a lower error in some cases.
We see this by comparing the dotted pink curve with the solid pink curve in the lower left panel.

The fourth panel shows sensitivity to the step-size parameter for HTD and TDRC.
Notice that compared to other Gradient-TD algorithms, TDRC has one sensitivity curve, shown in dashed blue.
This is because $\eta$ is set to one (also its regularization parameter was set to one) as proposed in the original paper.
HTD's widest curve was with $\eta=0.0625$ which was as wide as TDRC's curve.
For a more in-depth study of TDRC's extra parameters see Appendix~\ref{subapp:AdditionalResultsOfGradientTDAlgorithms}.

Among the Gradient-TD algorithms, TDRC was the easiest to use.
On the other hand, in the case of full bootstrapping, Proximal GTD2 reached the lowest error level among all Gradient-TD algorithms.
The fact that proximal GTD2 converged to a lower error level might be due to a few different reasons.
One possible reason is that it did not converge to the minimum of the mean squared projected Bellman error like other Gradient-TD methods.
Another reason might be that it converged to a minimum of the projected Bellman error that was different from the minimum the other algorithms converged to.
Further analyses is required to investigate this.
Overall, TDRC seems to be the easiest to use, but Proximal GTD2 achieves the lowest error on the Collision task.
It remains to be seen how these algorithms compare on future problems.

\section{Limitations and Future Work}
\label{sct:ConclusionsAndLimitations}

The present study is based on a single task, and this limits the conclusions that can be fairly drawn from it. For example, we have found that Emphatic-TD methods perform well over a wider range of parameters than Gradient-TD methods on the collision task, but it is entirely possible that the reverse would be true on a different task. Many more tasks must be explored before it is even possible for a consistent pattern to emerge that favors one class of algorithm over another. 

On the other hand, a pattern over empirical results must begin somewhere. We would stress the need for extensive empirical results even for a single task. Ours is the first systematic study of off-policy learning to describe the effects of all algorithm parameters (rather than, for example, simply taking the best performing parameters). Such a thorough examination is necessary to obtain the understanding that is critical to using off-policy algorithms successfully and with confidence. There is a need for thorough empirical studies, but they take time, and a proper presentation of them takes space. While our study is not the last word, it does contribute to the growing database of reliable results comparing modern off-policy learning algorithms.

Conducting additional experiments with other off-policy learning problems is a valuable direction for future work. In looking for the next problem, one might seek a task with greater challenges due to variance of the importance sampling ratios. In the Collision task, the product of ratios can grow nearly as large as $2^4= 16$. This could be made more extreme simply by increasing the number of states, or by changing the behavior policy. Also valuable would be exploring unrelated tasks with a different rationale for relevance to the real world.
One possibility is to use a task related to parallel learning about multiple alternative ways of behaving, such as learning how to exit each room in a four-rooms gridworld (Sutton, Precup \& Singh, 1999).

\begin{ack}
This work was funded by DeepMind, the Alberta Machine Intelligence Institute, NSERC, and CIFAR.
We would like to thank Ali Khlili Yegane for help in preparing the source code.
We thank Martha White, Adam White and Banafsheh Rafiee for useful feedback throughout the course of this project.
The computational resources of Compute Canada were essential to conducting this research.
\end{ack}

\section*{References}

\begin{list}{}{%
\setlength{\topsep}{0pt}%
\setlength{\leftmargin}{0.2in}%
\setlength{\listparindent}{-0.2in}%
\setlength{\itemindent}{-0.2in}%
\setlength{\parsep}{\parskip}%
}%
{ \small
\item[]
Baird, L. C. (1995). Residual algorithms: Reinforcement learning with function approximation. In \textit{Proceedings of the 12th International Conference on Machine Learning}, pp.~30--37.

\item[]
Baird, L. C. (1999). \textit{Reinforcement Learning through Gradient Descent.} PhD thesis, Carnegie
Mellon University.

\item[]
Boyan, J. A. (1999). Least-squares temporal difference learning. In \textit{Proceedings of the 16th
International Conference on Machine Learning,} pp.~49--56.

\item[]
Bradtke, S. J., Barto, A. G. (1996). Linear least-squares algorithms for temporal difference
learning. \textit{Machine Learning, 22} pp.~33--57.

\item[]
Dann, C., Neumann, G., Peters, J. (2014). Policy evaluation with temporal-differences: A survey and comparison. \textit{Journal of Machine Learning Research, 15} pp.~809--883.

\item[]
Espeholt, L., Soyer, H., Munos, R., Simonyan, K., Mnih, V., Ward, T., Doron, Y., Firoiu, V., Harley, T., Dunning, I. and Legg, S. (2018) IMPALA: Scalable distributed Deep-RL with importance weighted actor-learner architectures. In \textit{Proceedings of the 35th International Conference on Machine Learning.} pp.~1407--1416.

\item[]
Geist, M., Scherrer, B. (2014). Off-policy learning with eligibility traces: A survey. \textit{Journal
of Machine Learning Research 15} pp, 289--333.

\item[]
Ghiassian, S., Rafiee, B., Sutton, R. S. (2016). A first empirical study of emphatic temporal difference learning. In \textit{Workshop on Continual Learning and Deep Learning at the Conference on Neural Information Processing Systems.} ArXiv: 1705.04185.

\item[]
Ghiassian, S., Patterson, A., Garg, S., Gupta, D., White, A., White, M. (2020). Gradient temporal-difference learning with regularized corrections. In \textit{Proceedings of the 37th International Conference on Machine Learning,} pp.~3524--3534.

\item[]
Hackman, L. (2012). \textit{Faster Gradient-TD Algorithms.} MSc thesis, University of Alberta.

\item[]
Hallak, A., Tamar, A., Munos, R., Mannor, S. (2016). Generalized emphatic temporal-difference learning: Bias-variance analysis. In \textit{Proceedings of the Thirtieth AAAI Conference on Artificial Intelligence}, pp.~1631--1637.

\item[]
Jaderberg, M., Mnih, V., Czarnecki, W. M., Schaul, T., Leibo, J. Z., Silver, D., Kavukcuoglu, K. (2016). Reinforcement learning with unsupervised auxiliary tasks. ArXiv: 1611.05397.
}

\item[]
Juditsky, A., Nemirovski, A., Tauvel, C. (2011). Solving variational inequalities with stochastic mirror-prox algorithm. \textit{Stochastic Systems 1,} pp.~17--58.

\item[]
Liu B, Liu J, Ghavamzadeh M, Mahadevan S, Petrik M (2015). Finite-Sample Analysis of Proximal Gradient TD Algorithms. In \textit{Proceedings of the 31st International Conference on Uncertainty in Artificial Intelligence}, pp.~504--513.

\item[]
Liu B, Liu J, Ghavamzadeh M, Mahadevan S, Petrik M (2016). Proximal Gradient Temporal-Difference Learning Algorithms. In \textit{Proceedings of the 25th International Conference on Artificial Intelligence (IJCAI-16)}, pp.~4195--4199.

\item[]
Littman, M. L., Sutton, R. S., Singh, S. (2002). Predictive representations of state. In \textit{Advances in Neural Information Processing Systems 14,} pp.~1555--1561.

\item[]
Mahadevan, S., Liu, B., Thomas, P., Dabney, W., Giguere, S., Jacek, N., Gemp, I., Liu, J. (2014). Proximal reinforcement learning: A new theory of sequential decision making in primal--dual spaces. ArXiv: 1405.6757.

\item[]
Mahmood, A. R., Yu, H., Sutton, R. S. (2017). Multi-step off-policy learning without importance sampling ratios. ArXiv: 1702.03006.

\item[]
Maei, H. R. (2011). \textit{Gradient temporal-difference learning algorithms.} PhD thesis, University of Alberta.

\item[]
Modayil, J., Sutton, R. S. (2014). Prediction driven behavior: Learning predictions that drive fixed responses. In \textit{AAAI-14 Workshop on Artificial Intelligence and Robotics}.

\item[]
Munos, R., Stepleton, T., Harutyunyan, A., Bellemare, M. (2016). Safe and efficient off-policy reinforcement learning. In \textit{Advances in Neural Information Processing Systems 29,} pp.~1046--1054.

\item[]
Precup, D., Sutton, R. S., Dasgupta, S. (2001). Off-policy temporal-difference learning with function approximation. In \textit{Proceedings of the 18th International Conference on Machine Learning}, pp.~417--424.

\item[]
Precup, D., Sutton, R. S., Singh, S. (2000). Eligibility traces for off-policy policy evaluation. In \textit{Proceedings of the 17th International Conference on Machine Learning}, pp.~759--766.

\item[]
Rafiee, B., Ghiassian, S., White, A., Sutton, R. S. (2019). Prediction in Intelligence: An Empirical Comparison of Off-policy Algorithms on Robots. \textit{In Proceedings of the 18th International Conference on Autonomous Agents and MultiAgent Systems,} pp.~332--340.

\item[]
Ring, M. B. (in preparation). Representing knowledge as forecasts (and state as knowledge).

\item[]
Sutton, R. S. (1988). Learning to predict by the algorithms of temporal-differences. \textit{Machine Learning, 3} pp.~9--44.

\item[]
Sutton, R. S., Barto, A. G. (2018). \textit{Reinforcement Learning: An Introduction,} second edition. MIT press.

\item[]
Sutton, R. S., Maei, H. R., Precup, D., Bhatnagar, S., Silver, D., Szepesv\'ari, Cs., Wiewiora, E. (2009). Fast gradient-descent algorithms for temporal-difference learning with linear function approximation. In \textit{Proceedings of the 26th International Conference on Machine Learning}, pp.~993--1000.

\item[]
Sutton, R. S., Maei, H. R., Szepesvári, C. (2008). A convergent O(n) algorithm for off-policy temporal-difference learning with linear function approximation. In \textit{Advances in neural information processing systems 21,} pp.~1609--1616.

\item[]
Sutton, R. S., Mahmood, A. R., White, M. (2016). An emphatic approach to the problem of off-policy temporal- difference learning. \textit{Journal of Machine Learning Research, 17} pp.~1--29.

\item[]
Sutton, R. S., Modayil, J., Delp, M., Degris, T., Pilarski, P. M., White, A., Precup, D. (2011). Horde: A scalable real-time architecture for learning knowledge from unsupervised sensorimotor interaction. In \textit{Proceedings of the 10th International Conference on Autonomous Agents and Multiagent Systems}, pp.~761--768.

\item[]
Sutton, R. S., Precup, D., Singh, S. (1999). Between MDPs and semi-MDPs: A framework for temporal abstraction in reinforcement learning. \textit{Artificial Intelligence, 112} pp.~181--211.

\item[]
Tanner, B., Sutton, R. S., (2005). TD\la networks: temporal-difference networks with eligibility traces. In \textit{Proceedings of the 22nd international conference on Machine learning,} pp.~888--895.

\item[]
Thomas, P. S. (2015). \textit{Safe reinforcement learning.} PhD thesis, University of Massachusetts Amherst.

\item[]
Touati, A., Bacon, P. L., Precup, D., Vincent, P. (2017). Convergent tree-backup and retrace with function approximation. ArXiv: 1705.09322.

\item[]
Watkins, C. J. C. H. (1989). \textit{Learning from delayed rewards.} PhD thesis, University of Cambridge.

\item[]
Watkins, C. J. C. H., Dayan, P. (1992). Q-learning. \textit{Machine Learning, 8} pp.~279--292.

\item[]
White, A. (2015). \textit{Developing a predictive approach to knowledge.} PhD thesis, University of
Alberta.

\item[]
White, M. (2017). Unifying Task Specification in Reinforcement Learning. In \textit{Proceedings of the 34th International Conference on Machine Learning,} pp.~3742--3750.

\item[]
White, A., White, M. (2016). Investigating practical linear temporal difference learning. In
\textit{Proceedings of the 2016 International Conference on Autonomous Agents and Multiagent Systems,} pp.~494--502.

\item[]
Yu, H., Mahmood, A. R., Sutton, R. S. (2018). On generalized bellman equations and temporal-difference learning. \textit{The Journal of Machine Learning Research, 19(1),} pp.~1864--1912.
\end{list}

\appendix

\newpage

\section{Update Rules}
\label{app:PseudoCodes}

In this section we list the update rules for all algorithms empirically studied in the text.
This list provides a concise reference for the update rules for each algorithm.
We include this list as a single point of reference for each algorithm.
\newline

\noindent\textbf{TD\la:}
\begin{align*}
\delta_t \defeq&\,\, R_{t+1} + \gamma_{t+1} \vecw_{t}^{\tr}\vecx_{t+1} - \vecw_{t}^{\tr}\vecx_{t}\\
\vecz_t \leftarrow& ~\rho_t(\gamma_t \lambda_t \vecz_{t-1}  +\vecx_t) \textnormal{\qquad with } \vecz_{-1} = \bf{0}\nonumber\\
\vecw_{t+1} \leftarrow& ~ \vecw_t +\alpha \delta_t \vecz_t
\end{align*}

\noindent\textbf{GTD\la:}
\begin{align*}
\delta_t \defeq&\,\, R_{t+1} + \gamma_{t+1}\vecw_t^{\tr}\vecx_{t+1} - \vecw_t^{\tr}\vecx_{t}\nonumber\\
\vecz_t \leftarrow& ~\rho_t(\gamma_t \lambda_t \vecz_{t-1}  +\vecx_t) \textnormal{\qquad with } \vecz_{-1} = \bf{0}\nonumber\\
\vecv_{t+1} \leftarrow& ~\vecv_t + \alpha_\vecv\biggl[\delta_t\vecz_t - (\vecv_t^{\tr}\vecx_t)\vecx_{t}  \biggr] \nonumber\\\
\vecw_{t+1} \leftarrow& ~\vecw_t + \alpha \delta_t \vecz_t - \alpha \gamma_{t+1} (1-\lambda_{t+1})(\vecv_t^{\tr}\vecz_t)\vecx_{t+1}
\end{align*}

\noindent\textbf{TDRC\la:}
\begin{align*}
\delta_t \defeq&\,\, R_{t+1} + \gamma_{t+1}\vecw_t^{\tr}\vecx_{t+1} - \vecw_t^{\tr}\vecx_{t}\nonumber\\
\vecz_t \leftarrow& ~\rho_t(\gamma_t \lambda_t \vecz_{t-1}  +\vecx_t) \textnormal{\qquad with } \vecz_{-1} = \bf{0}\nonumber\\
\vecv_{t+1} \leftarrow& ~\vecv_t + \alpha\biggl[\delta_t\vecz_t - (\vecv_t^{\tr}\vecx_t)\vecx_{t}  \biggr] - \alpha\vecv_t \nonumber\\\
\vecw_{t+1} \leftarrow& ~\vecw_t + \alpha \delta_t \vecz_t - \alpha \gamma_{t+1} (1-\lambda_{t+1})(\vecv_t^{\tr}\vecz_t)\vecx_{t+1}
\end{align*}

\noindent\textbf{GTD2\la:}
\begin{align*}
\delta_t \defeq&\,\, R_{t+1} + \gamma_{t+1}\vecw_t^{\tr}\vecx_{t+1} - \vecw_t^{\tr}\vecx_{t}\nonumber\\
\vecz_t \leftarrow& ~\rho_t(\gamma_t \lambda_t \vecz_{t-1}  +\vecx_t) \textnormal{\qquad with } \vecz_{-1} = \bf{0}\nonumber\\
\vecv_{t+1} \leftarrow& ~\vecv_t + \alpha_\vecv\biggl[\delta_t\vecz_t - (\vecv_t^{\tr}\vecx_t)\vecx_{t}  \biggr] \nonumber\\
\vecw_{t+1} \leftarrow& ~\vecw_t +\alpha (\vecv_t^{\tr}\vecx_t)\vecx_t - \alpha \gamma_{t+1} (1 - \lambda_{t+1}) (\vecv_t^{\tr}\vecz_t)\vecx_{t+1}
\end{align*}

\noindent\textbf{HTD\la:}
\begin{align*}
\delta_t \defeq&\,\, R_{t+1} + \gamma_{t+1}\vecw_t^{\tr}\vecx_{t+1} - \vecw_t^{\tr}\vecx_{t}\nonumber\\
\vecz_t^\rho \leftarrow& ~\rho_t(\gamma \lambda \vecz_{t-1}^{\rho}  +\vecx_t) \textnormal{\qquad with } \vecz_{-1} = \bf{0}\nonumber\\
\vecz_t \leftarrow& ~\gamma \lambda \vecz_{t-1}  +\vecx_t\nonumber\\
\vecv_{t+1} \leftarrow& ~\vecv_t + \alpha_\vecv\biggl[\delta_t\vecz_t - (\vecx_t - \gamma_{t+1}\vecx_{t+1})(\vecv_t^{\tr}\vecz_t)  \biggr] \nonumber\\\
\vecw_{t+1} \leftarrow& ~\vecw_t + \alpha\biggl[\delta_t\vecz_t + (\vecx_t - \gamma_{t+1}\vecx_{t+1})(\vecz_t -\vecz_t)^{\tr}\vecv_t  \biggr]
\end{align*}

\noindent\textbf{Proximal GTD2\la:}
\begin{align*}
\delta_t \defeq&\,\, R_{t+1} + \gamma_{t+1}\vecw_t^{\tr}\vecx_{t+1} - \vecw_t^{\tr}\vecx_{t}\nonumber\\
\vecz_t \leftarrow& ~\rho_t(\gamma_t \lambda_t \vecz_{t-1}  +\vecx_t)\textnormal{\quad with } \vecz_{-1} = \bf{0}\nonumber\\
\vecv_{t+\frac{1}{2}} \leftarrow& ~\vecv_{t} + \alpha_\vecv \biggl[\delta_{t} \vecz_t - (\vecv_{t}^{\tr}\vecx_t)\vecx_t\biggr]\\
\vecw_{t+\frac{1}{2}} \leftarrow& ~\vecw_t +\alpha (\vecv_t^{\tr}\vecx_t)\vecx_t -\alpha \gamma_{t+1} (1 - \lambda_{t+1}) (\vecv_t^{\tr}\vecz_{t}^{\rho} ) \vecx_{t+1}\\
\delta_{t+\frac{1}{2}} \defeq&\,\, R_{t+1} + \gamma_{t+1} \vecw_{t+\frac{1}{2}}^{\tr}\vecx_{t+1} - \vecw_{t+\frac{1}{2}}^{\tr}\vecx_{t}\\
\vecv_{t+1} \leftarrow& ~\vecv_{t} + \alpha_\vecv \biggl[\delta_{t+\frac{1}{2}} \vecz_t^\rho - (\vecv_{t+\frac{1}{2}}^{\tr}\vecx_t)\vecx_t\biggr]\\
\vecw_{t+1} \leftarrow& ~\vecw_t +\alpha (\vecv_{t+\frac{1}{2}}^{\tr}\vecx_t)\vecx_t -\alpha  \gamma_{t+1} (1 - \lambda_{t+1}) (\vecv_{t+\frac{1}{2}}^{\tr}\vecz_{t}^{\rho} ) \vecx_{t+1}
\end{align*}

\noindent\textbf{Emphatic TD\la:}
\begin{align*}
\delta_t \defeq&\,\, R_{t+1} + \gamma_{t+1}\vecw_t^{\tr}\vecx_{t+1} - \vecw_t^{\tr}\vecx_{t}\nonumber\\
F_t \leftarrow& ~\rho_{t-1}\gamma_t F_{t-1} + I_t \textnormal{\quad with } F_{0} = I_0\\
M_{t}  ~\defeq&\,\, \lambda_t I_t + (1 - \lambda_t)F_t\\
\vecz_t \leftarrow& ~\rho_t \left(\gamma_t \lambda \vecz_{t-1} + M_t \vecx_{t}\right)\textnormal{\quad with } \vecz_{-1} = \bf{0}\\
\vecw_{t+1} \leftarrow& ~ \vecw_t +\alpha \delta_t \vecz_t
\end{align*}

\noindent\textbf{Emphatic TD($\lambda\, ,\beta$):}
\begin{align*}
\delta_t \defeq&\,\, R_{t+1} + \gamma_{t+1}\vecw_t^{\tr}\vecx_{t+1} - \vecw_t^{\tr}\vecx_{t}\nonumber\\
F_t \leftarrow& ~\rho_{t-1}\beta F_{t-1} + I_t \textnormal{\quad with } F_{0} = I_0\\
M_{t}  \defeq&\,\, \lambda_t I_t + (1 - \lambda_t)F_t\\
\vecz_t \leftarrow& ~\rho_t \left(\gamma_t \lambda \vecz_{t-1}^{\rho} + M_t \vecx_{t}\right)\textnormal{\quad with } \vecz_{-1} = \bf{0}\\
\vecw_{t+1} \leftarrow& ~ \vecw_t +\alpha \delta_t \vecz_t
\end{align*}

\noindent\textbf{Tree Backup\la for prediction:}
\begin{align*}
\delta_t^{\rho} \defeq&\,\, \rho_t \biggl( R_{t+1} + \gamma_{t+1}\vecw_t^{\tr}\vecx_{t+1} - \vecw_t^{\tr}\vecx_{t}\biggr)\nonumber\\
\vecz_t \leftarrow& ~\gamma_t \lambda_t \pi_{t-1} \vecz_{t-1} + \vecx_{t}\textnormal{\quad with } \vecz_{-1} = \bf{0}\\
\vecw_{t+1} \leftarrow& ~\vecw_t +\alpha \delta_t^{\rho} \vecz_t
\end{align*}

\noindent\textbf{Vtrace\la:}
\begin{align*}
\delta_t \defeq&\,\, R_{t+1} + \gamma_{t+1}\vecw_t^{\tr}\vecx_{t+1} - \vecw_t^{\tr}\vecx_{t}\nonumber\\
\vecz_t \leftarrow& ~\textnormal{max}(\rho_t, 1) \left(\gamma_t \lambda \vecz_{t-1} + \vecx_{t}\right)\textnormal{\quad with } \vecz_{-1} = \bf{0}\\
\vecw_{t+1} \leftarrow& ~\vecw_t +\alpha \delta_t \vecz_t
\end{align*}

\noindent\textbf{ABTD($\zeta$):}
\begin{align*}
\delta_t^{\rho} \defeq&\,\, \rho_t \biggl( R_{t+1} + \gamma_{t+1}\vecw_t^{\tr}\vecx_{t+1} - \vecw_t^{\tr}\vecx_{t}\biggr)\nonumber\\
\vecz_t \leftarrow& ~\gamma_t \nu_{t-1} \pi_{t-1} \vecz_{t-1} + \vecx_{t}\textnormal{\quad with } \vecz_{-1} = \bf{0}\\
\vecw_{t+1} \leftarrow& ~ \vecw_t +\alpha \delta_t^{\rho} \vecz_t
\end{align*}

\newpage

\section{Parameter Settings}
\label{app:ParameterSettings}

Parameter settings for all algorithms are listed in Table~\ref{tab:params-tbl}.
All algorithms used the same set of primary step sizes and the same set of bootstrapping parameters.
For algorithms that had an extra parameter, such as GTD\la, we tried all combinations of the first step size, bootstrapping parameter, and the extra parameter as listed in the table.

\begin{table}[h!]
\caption{List of all parameters used in the experiment.}
\begin{tabular}{|c|c|c|c|c|}
\hline
\multicolumn{2}{|c|}{\textbf{\begin{tabular}[c]{@{}c@{}}$~$\\ Algorithms\\ $~$\end{tabular}}} & \textbf{$\eta$ or $\beta$} & \textbf{$\lambda$ or $\zeta$} & \textbf{$\alpha$} \\ \hline
\multicolumn{2}{|c|}{\begin{tabular}[c]{@{}c@{}}$~$\\ Off-policy TD($\lambda$)\\ $~$\end{tabular}} & --- & \multirow{11}{*}{\begin{tabular}[c]{@{}c@{}}0, 0.1, 0.2, \\ 0.3, 0.5, 0.9, 1\\ and\\ 1 - $2^{-x}$\\ where\\ $x \in \{$\\ $2, 3, 4, 5, 6\}$\end{tabular}} & \multirow{11}{*}{\begin{tabular}[c]{@{}c@{}}$\alpha = 2^{-x}$\\ where\\ $x \in \{$\\ $0, 1, 2, \cdots, $\\ $17, 18\}$\end{tabular}} \\ \cline{1-3}
\multirow{5}{*}{\begin{tabular}[c]{@{}c@{}}Gradient-TD\\ Algorithms\end{tabular}} & \begin{tabular}[c]{@{}c@{}}$~$\\ GTD($\lambda$)\\ $~$\end{tabular} & \multirow{4}{*}{\begin{tabular}[c]{@{}c@{}}$2^{x}$\\ where\\ $x \in \{-6, -5, \cdots, 7, 8 \}$\end{tabular}} &  &  \\ \cline{2-2}
 & \begin{tabular}[c]{@{}c@{}}$~$\\ GTD2($\lambda$)\\ $~$\end{tabular} &  &  &  \\ \cline{2-2}
 & \begin{tabular}[c]{@{}c@{}}$~$\\ HTD($\lambda$)\\ $~$\end{tabular} &  &  &  \\ \cline{2-2}
 & \begin{tabular}[c]{@{}c@{}}$~$\\ Proximal GTD2($\lambda$)\\ $~$\end{tabular} &  &  &  \\ \cline{2-3}
 & \begin{tabular}[c]{@{}c@{}}$~$\\ TDRC($\lambda$)\\ $~$\end{tabular} & --- &  &  \\ \cline{1-3}
\multirow{2}{*}{\begin{tabular}[c]{@{}c@{}}Emphatic-TD\\ Algorithms\end{tabular}} & \begin{tabular}[c]{@{}c@{}}$~$\\ Emphatic\\ TD($\lambda$)\\ $~$\end{tabular} & --- &  &  \\ \cline{2-3}
 & \begin{tabular}[c]{@{}c@{}}$~$\\ Emphatic\\ TD($\lambda,~\beta$)\\ $~$\end{tabular} & \begin{tabular}[c]{@{}c@{}}$\beta\in$\\ \{0.0, 0.2, 0.4, 0.6, 0.8, 1.0\}\end{tabular} &  &  \\ \cline{1-3}
\multirow{3}{*}{\begin{tabular}[c]{@{}c@{}}Variable-$\lambda$\\ Algorithms\end{tabular}} & \begin{tabular}[c]{@{}c@{}}$~$\\ Tree Backup($\lambda$)\\ $~$\end{tabular} & \multirow{3}{*}{---} &  &  \\ \cline{2-2}
 & \begin{tabular}[c]{@{}c@{}}$~$\\ Vtrace($\lambda$)\\ $~$\end{tabular} &  &  &  \\ \cline{2-2}
 & \begin{tabular}[c]{@{}c@{}}$~$\\ ABTD($\zeta$)\\ $~$\end{tabular} &  &  &  \\ \hline
\end{tabular}
\label{tab:params-tbl}
\end{table}

\newpage

\section{Off-policy Temporal-Difference Algorithms for Prediction Learning}
\label{app:TemporalDifferenceAlgorithmsForOffPolicyPrediction}

In this section, we explain the algorithms we used in the experiments in more detail.
We start with the Off-policy TD\la algorithm.
By showing how TD\la (Sutton, 1988) diverges on a simple problem with two states, we motivate the rest of the algorithms that are guaranteed convergent under off-policy training, including gradient and emphatic families of algorithms.


\subsection[The Off-policy TD Algorithm]{The Off-policy TD\la Algorithm}
\label{subapp:OffPolicyTD0}
The value of a state $S_t=s$ can be written recursively, using the value of the next state $S_{t+1}$.
To show this, we use the following identity:
\begin{align}
\E{X}=\E{\CE{X}{Y}}, \label{eq:TowerRule}
\end{align}
which in statistics, is known as the law of total expectation or the tower rule.
We also know that:
\begin{align}
v_\pi(s) \defeq \CEpi{G_t}{S_t=s}, \nonumber 
\end{align}
and we have the same definition for the next state, $s'$:
\begin{align}
v_\pi(s') \defeq \CEpi{G_{t+1}}{S_{t+1}=s'}. \label{eq:DefinitionOfVPiPrime}
\end{align}
Using (\ref{eq:TowerRule}) and (\ref{eq:DefinitionOfVPiPrime}), the value of the next state $S_{t+1}$ can be written:
\begin{align}
\CEpi{v_\pi(S_{t+1})}{S_t=s} &= \CEpi{\CEpi{G_{t+1}}{S_{t+1}=s'}}{S_t=s}, \nonumber\\
&=\CEpi{G_{t+1}}{S_t=s}, \label{eq:v_s_t+1IsEqualtoEG_t+1} 
\end{align}
which is used below to show that the value of the current state $S_t=s$ is equal to the expectation of the reward plus the discounted value of the next state:
\begin{align}
v_\pi(s) & \defeq \CEpi{G_t}{S_t=s}, \nonumber\\
& = \CEpi{R_{t+1} + \gamma G_{t+1}}{S_t=s}, \nonumber\\
& = \CEpi{R_{t+1}}{S_t=s} + \gamma \CEpi{G_{t+1}}{S_t=s},\tag{by (\ref{eq:v_s_t+1IsEqualtoEG_t+1})}\\
& = \CEpi{R_{t+1} + \gamma v_\pi(S_{t+1})}{S_t=s}.\label{eq:BellmanEq}  
\end{align}
Equation (\ref{eq:BellmanEq}) is the \emph{Bellman equation} for $v_\pi$. 
The Bellman equation says that the value of a state is equal to the expectation of the reward plus the value of the next state.
If $v_\pi(s)$ is moved to the right hand side and inside the expectation (it can be moved inside the expectation because it is a constant), we get:
\begin{align}
\CEpi{R_{t+1} + \gamma v_\pi(S_{t+1}) - v_\pi(s)}{S_t=s} = 0, \label{eq:BellmanError}
\end{align}
which is referred to as the \emph{Bellman error}. 
The error is equal to 0 for the true value function, $v_\pi$.

Temporal-difference learning updates the weight vector, $\vecw$, in the direction that minimizes the Bellman error.
Note that~(\ref{eq:BellmanError}) holds only approximately when using function approximation:
\begin{align}
&\CEpi{R_{t+1} + \gamma \vhat_\vecw(S_{t+1}) - \vhat_\vecw(s)}{S_t=s} \approx 0. \label{eq:ApproximateBellmanError}
\end{align}
The simplest form of temporal-difference learning, the TD(0) algorithm, uses a sample of the left hand side of (\ref{eq:ApproximateBellmanError}) to update $\vecw$. This sample is called the \emph{TD error} and is defined as:
\begin{align}
&\delta_t  \defeq R_{t+1} + \gamma \vecw_{t}\tr\vecx_{t+1} - \vecw_{t}\tr\vecx_{t}. \label{eq:delta}
\end{align}
Finally, TD(0) uses (\ref{eq:delta}) to update the weight vector as follows:
\begin{align}
\vecw_{t+1} & \leftarrow \alpha \delta_t \vecx_{t}, \label{eq:on_policy_td_0}
\end{align}
where $\alpha$ is a scalar constant step-size parameter at time-step.
If $\alpha$ is adaptive and changes at each time step, it is denoted by $\alpha_t$.

In off-policy learning, the agent follows a behaviour policy $b$, but learns the value function for a different target policy $\pi$. 
To account for the difference between the target and behaviour policies, importance sampling ratios are typically used.
The importance sampling ratio is the probability of taking an action in a state under the target policy divided by the same probability under the behaviour policy. Formally:
\begin{align}
\rho_t \defeq \frac{\pi(A_t|S_t)}{b(A_t|S_t)}.
\end{align}

\emph{Off-policy TD(0)} augments the TD(0) update, (\ref{eq:on_policy_td_0}), with an importance sampling ratio:
\begin{align}
\vecw_{t+1} & \leftarrow \vecw_{t} + {\rho_t} \alpha \delta_t \vecx_{t}. \label{eq:off_policy_td_0}
\end{align}

One of the widely known ideas to make TD-style algorithms more efficient, are \emph{eligibility traces}. Eligibility traces help assign credit to features that were activated in the past based on how far in the past they were activated. They assign credit by storing a fading trace of the features that were activated. 
We focus on one form of eligibility trace called the accumulating trace. 
The Off-policy TD\la algorithm with eligibility traces is thoroughly explained by the following update rules:
\begin{align}
\vecz_{t} & \leftarrow \vecz_{t-1} + \rho_t (\gamma \lambda \vecz_{t-1} + \vecx_t), \textnormal{\qquad with } \vecz_{-1} = \bf{0} \nonumber, \\
\vecw_{t+1} & \leftarrow \vecw_{t} +  \alpha \delta_t \vecz_{t}. \nonumber 
\end{align}

Off-policy TD\la might diverge when combined with linear function approximation.
A simple example can help see this intuitively (Sutton \& Barto, 2018).
Suppose, two states in an MDP, whose values are approximated using the same weight $w$, except that the second state has a value twice as big as the first state as shown below\footnote{The weight vector in this example is a single number represented by $w$, and the feature vectors are $1$ for the left state and $2$ for the right state.}.
Suppose, for this MDP that $\gamma=1$ and the rewards are $0$ on all transitions.

If the transition from $w$ to $2w$ is experienced repeatedly, the weight will diverge to infinity. 
Suppose, the first time the transition is experienced, $w=10$.
This means $2w$ is $20$ and the TD error will be $10$.
If, for example, $\alpha=0.5$, the value of $w$ will be increased by $5$ and will change to $15$.
If this transition is experienced repeatedly, it is clear that the weights will diverge to infinity.
\begin{center}
\begin{tikzpicture}[
roundnode/.style={circle, draw=black, very thick, minimum size=10mm},
]
\node[roundnode]      (maintopic)                              {$w$};
\node[roundnode]      (rightsquare)       [right=of maintopic] {$2w$};

\draw[-|>] (maintopic.east) -- (rightsquare.west);
\end{tikzpicture}
\end{center}

There is no way on-policy TD(0) may diverge in the example above. In on-policy learning, when the agent moves from $2w$ to $w$, through some path (not shown in the figure above), the weights will eventually decrease and convergence will be guaranteed. 
In off-policy learning, however, there can be situations where the weights do not decrease when taking the path from $2w$ to $w$ and divergence will occur.


\subsection{Gradient-TD Algorithms}
\label{subapp:GradientTDAlgorithms}

One way to assure convergence is to use stochastic gradient descent to minimize an objective.
By using stochastic gradient descent, desirable theoretical guarantees including convergence will follow, even in the off-policy learning case.
One important question is: What objective function should be minimized?

\begin{figure}[t]
      \centering
      \includegraphics[width=0.4\linewidth]{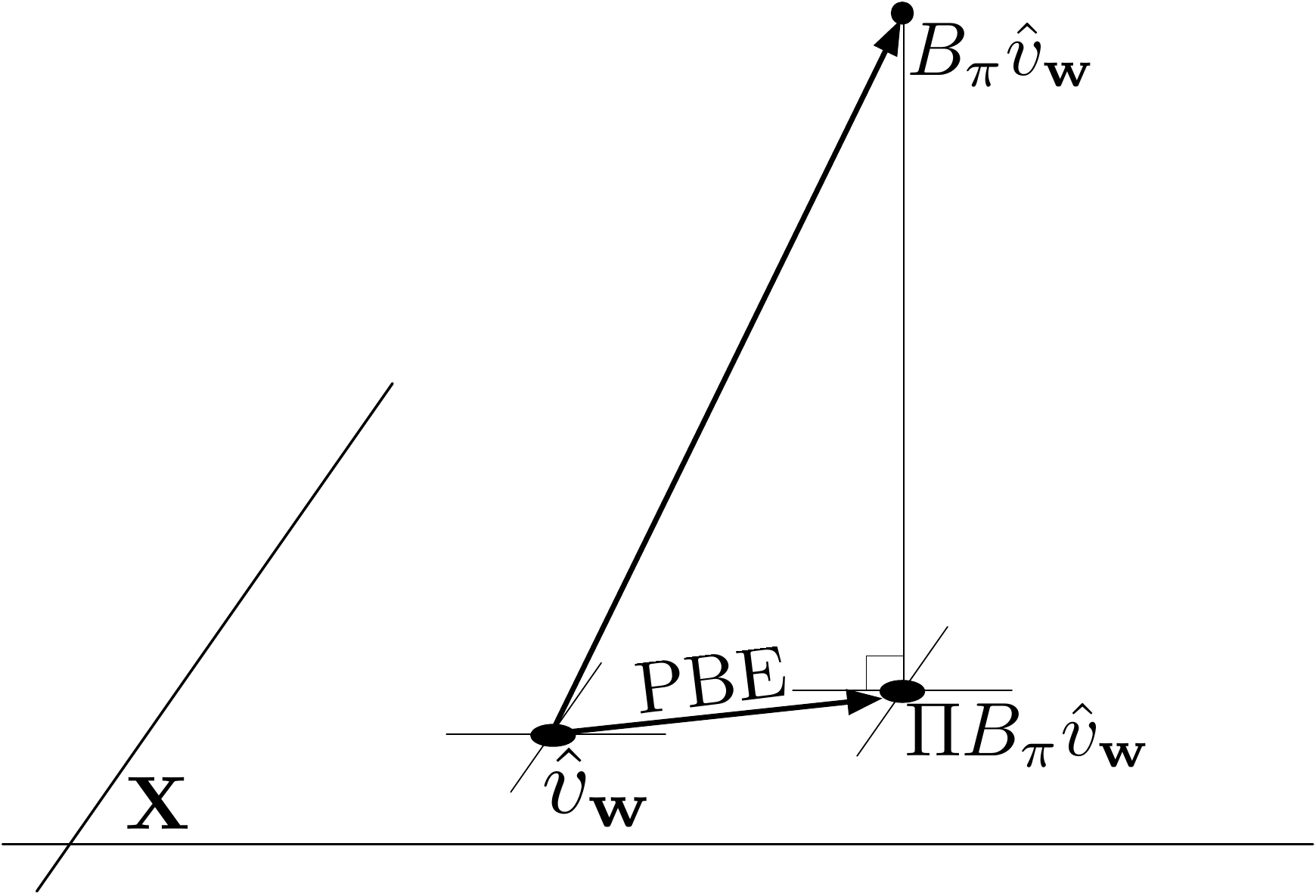}
      \caption{Geometry of linear function approximation shown in three dimensional space.}
      \label{fig:Value_function_geometry}
\end{figure}

One option is to minimize a the mean squared projected Bellman error, which we denote by $\PBE$.
To define $\PBE$, we first need to define the Bellman operator.
The Bellman error, in state $s$, can be written:
\begin{align}
& \CEpi{R_{t+1}+\gamma \hat{v}_{\vecw}(S_{t+1}) - \hat{v}_{\vecw}(S_{t})}{S_t=s}, \nonumber
\intertext{or in explicit form:}
& \bigg(\sum_{a}\pi(a|s)\sum_{s', r}p(s', r|s, a)[r + \gamma \vhat_{\vecw}(s')] \bigg) - \vhat_\vecw(s), \label{eq:bellman_error_flushed}
\end{align}
where $p$ is the underlying transition probability distribution of the MDP.
 The Bellman operator $B_\pi:\mathbb{R}^{|\mathcal{S}|}\rightarrow\mathbb{R}^{|\mathcal{S}|}$, at state $s$, is defined as:
\begin{align}
(B_\pi \vhat_\vecw)(s) \defeq \sum_{a}\pi(a|s)\sum_{s', r}p(s', r|s, a)[r + \gamma \vhat_\vecw(s')]. \label{eq:bellman_operator}
\end{align}
As seen in \eqref{eq:bellman_operator}, the Bellman operator works irrespective of the parameters of the value function.
This means after applying the Bellman operator to a value function, the new value function might not be representable by in the feature space.
It can, however, be projected back into the space of the representable functions, using a projection matrix.
In the linear function approximation case, the projection operator is linear, meaning that it can be represented as an $|\mathcal{S}| \times |\mathcal{S}|$ matrix:
$$\Pi \defeq \Xmat (\Xmat\tr\Dmat\Xmat)^{-1}\Xmat\tr\Dmat,$$
where $\Xmat$ is a matrix of size $|\mathcal{S}| \times d$ with feature vectors of all states as its rows and $\Dmat$ is a diagonal matrix with the state visitation norm on its diagonal. 
To minimize the distance between the value function and the projection of the value function after applying the Bellman operator, we first need to apply the Bellman operator to the current value function as in (\ref{eq:bellman_operator}). 
The Bellman error for all states, put together in a vector is called the Bellman error vector. 
We denote the Bellman error vector by $\bar{\delta}_\vecw$ ((\ref{eq:bellman_error_flushed}) for all $s$). 
This vector can be projected to the space of representable value function.
We denote the projected vector by $\Pi\bar{\delta}_\vecw$. 
The distance between the value function before applying the Bellman operator and the projection of the function after applying the Bellman operator is known as the \emph{Mean Squared Projected Bellman Error} or the $\PBE$ and can be minimized directly using stochastic gradient descent.
See Figure~\ref{fig:Value_function_geometry}. 
The $\PBE$ is defined as:
\begin{align}
\PBE(\vecw) = || \Pi \bar{\delta}_\vecw ||_{\mu_b}^2 \label{eq:PBE},
\end{align}
where $\mu_b$ is the state visitation distribution induced by policy $b$.

All members of the Gradient-TD family minimize some form of the mean squared projected Bellman error. 
GTD\la and GTD2\la both minimize the $\PBE$ objective (Sutton et al. 2009). 
An earlier version of the Gradient-TD family minimizes the \emph{Norm of the Expected TD Update}, or the NEU (Sutton, Maei, \& Szepesvari, 2008). 
The NEU does not have a readily available geometric interpretation, but it is only different from the $\PBE$ in how the objective function (\ref{eq:PBE}) is weighted.\footnote{The algorithm proposed by Sutton, Maei, \& Szepesvari (2008) was called GTD. The algorithms proposed by Sutton et al. (2009) were called TDC and GTD2.
Later on, in Maei (2011), the TDC and GTD2 algorithms were combined with eligibility traces and were called GTD\la and GTD2\la, respectively; meaning that GTD(0) is the same as TDC and GTD2(0) is the same as GTD2.}
The algorithm that minimizes the NEU was later found to be inferior to GTD and GTD2 algorithms and was abandoned (Sutton et al., 2009; Dann, Neumann, \& Peters, 2014).

Update rules for GTD\la and GTD2\la are similar to TD\la.
The only difference is that the Gradient-TD algorithms have a correction term that changes the original TD update to make sure it follows the direction of the gradient of $\PBE$ at each time step.
To estimate the correction term, both algorithms use an additional learned weight vector, denote by $\vecv$, that can be learned at a different rate than the primary weight vector.
The update rules for GTD\la are:
\begin{align*}
\vecz_t \leftarrow& ~\rho_t(\gamma_t \lambda_t \vecz_{t-1}  +\vecx_t), \textnormal{\qquad with } \vecz_{-1} = \bf{0},\nonumber\\
\vecv_{t+1} \leftarrow& ~\vecv_t + \alpha_\vecv\biggl[\delta_t\vecz_t - (\vecv_t^{\tr}\vecx_t)\vecx_{t}  \biggr], \nonumber\\\
\vecw_{t+1} \leftarrow& ~\vecw_t + \alpha \delta_t \vecz_t - \underbrace{\alpha \gamma_{t+1} (1-\lambda_{t+1})(\vecv_t^{\tr}\vecz_t)\vecx_{t+1}}_{\text{correction term}},
\end{align*}
\noindent and the update rules for GTD2\la are:
\begin{align*}
\vecz_t \leftarrow& ~\rho_t(\gamma_t \lambda_t \vecz_{t-1}  +\vecx_t), \textnormal{\qquad with } \vecz_{-1} = \bf{0},\nonumber\\
\vecv_{t+1} \leftarrow& ~\vecv_t + \alpha_\vecv\biggl[\delta_t\vecz_t - (\vecv_t^{\tr}\vecx_t)\vecx_{t}  \biggr], \nonumber\\
\vecw_{t+1} \leftarrow& ~\vecw_t +\alpha (\vecv_t^{\tr}\vecx_t)\vecx_t - \alpha \gamma_{t+1} (1 - \lambda_{t+1}) (\vecv_t^{\tr}\vecz_t)\vecx_{t+1}.
\end{align*}

There were a few attempts at making GTD\la and GTD2\la faster.
HTD\la, was the first such algorithm.
Hybrid TD was first proposed by Maei (2011).
It was later developed more by Hackman (2012).
It was finally extended to the eligibility trace case by White \& White (2016).
Sutton et al. (2009) showed that although Gradient-TD algorithms provide better convergence guarantees than off-policy TD, they learn slower. 
HTD\la was an attempt at combining the fast learning of Off-policy TD\la and the convergence guarantees of the Gradient-TD family. 
HTD\la is a generalization of TD\la in the sense that in the on-policy case, HTD\la reduces to TD\la.
To better understand HTD\la, we first write the $\PBE$ in an alternative form:
\begin{align}
\PBE(\vecw)=\Eb{\delta_t\vecz_t}\tr \Eb{\vecx(S_t)\vecx(S_t)\tr}^{-1} \Eb{\delta_t\vecz_t} \nonumber,
\end{align}
\noindent We then use the following definitions:
\begin{align*}
\Cmat &\defeq \Eb{\vecx(S_t) \vecx(S_t)\tr},\\
\Amat &\defeq -\Eb{(\gamma \vecx(S_{t+1}) - \vecx(S_t)) \vecx(S_t)\tr},\\
\vecb &\defeq  \Eb{R(S_t,A_t) \vecx(S_t)\tr},
\end{align*}
\noindent where $R$ is the reward function and $\vecx$ is a function that takes in as input the state and returns the feature representation vector.
\noindent Using the above identities, we write the $\PBE$ in the following form (Hackman, 2012):
\begin{align}
\PBE &= (\sneg\Amat \vecw + \vecb)^\top \Cmat^\inv (\sneg\Amat \vecw + \vecb) \label{eq:PBEABC},
\intertext{where}
- \Amat \vecw + \vecb &= \Eb{\delta_t\vecz_t} \nonumber,\\
\Cmat &= \Eb{\vecx(S_t)\vecx(S_t)\tr} \nonumber.
\end{align}
In \eqref{eq:PBEABC}, the $\Cmat$ matrix is weighting $- \Amat \vecw + \vecb $. 
As long as $- \Amat \vecw + \vecb $ becomes $\vec0$ asymptotically, the algorithms find a solution to the $\PBE$ and the weighting, $\Cmat$ is irrelevant in the quality of the solution. 
However, the rate of convergence might change with $\Cmat$.
More specifically, the $\Cmat$ matrix can be replaced by any positive definite matrix and the resulting algorithm is guaranteed to be stable and will converge to the minimum of the $\PBE$. 
The HTD\la algorithm replaces $\Cmat^{-1}$ with $\Xmat^{-\top}$, where:
\begin{align}
\Xmat^{-\top} \defeq \Eb{\big(\vecx(S_t) - \gamma \vecx(S_{t+1})\big)\vecz_t^{\vecb\tr}} \nonumber,
\end{align}
\noindent where $\vecz_t^\vecb$ is an eligibility trace vector that does not use importance sampling ratios, or simply the on-policy trace:
\begin{align}
\vecz_t^\vecb \defeq \gamma \lambda \vecz_{t-1}^\vecb + \vecx_t. \label{eq:OnpolicyTrace}
\end{align}
Using $\Xmat^{-\top}$ instead of $\Cmat^{-1}$, and computing the derivative of the resulting $\PBE$, we get the HTD\la update rules (White \& White, 2016):
\begin{align*}
\vecz_t \leftarrow& ~\rho_t(\gamma \lambda \vecz_{t-1}  +\vecx_t), \textnormal{\qquad with } \vecz_{-1} = \bf{0},\\
\vecz_t^\vecb \leftarrow& ~\gamma \lambda \vecz_{t-1}^\vecb  +\vecx_t, \textnormal{\qquad with }  \vecz_{-1}^\vecb = \bf{0},\\
\vecv_{t+1} \leftarrow& ~\vecv_t + \alpha_\vecv\biggl[\delta_t\vecz_t - (\vecx_t - \gamma_{t+1}\vecx_{t+1})(\vecv_t^{\tr}\vecz_t^\vecb)  \biggr],\\
\vecw_{t+1} \leftarrow& ~\vecw_t + \alpha\biggl[\delta_t\vecz + (\vecx_t - \gamma_{t+1}\vecx_{t+1})(\vecz_t -\vecz_t^\vecb)^{\tr}\vecv_t  \biggr],
\end{align*}
\noindent where $\vecz_t$ is the normal off-policy trace with the importance sampling ratio and $\vecz_t^b$ is the on-policy trace.

Proximal Gradient-TD algorithms are another set of algorithms proposed to improve on the original Gradient-TD algorithms (Mahadevan et al., 2014; Liu et al., 2015; Liu et al., 2016). 
Proximal Gradient-TD algorithms improve on classic Gradient-TD algorithms in the sense that they use the true gradient of the $\PBE$ objective to update the weight vector.
This is in contrast to Gradient-TD algorithms that are not true stochastic gradient algorithms with respect to the $\PBE$ objective. 
The reason why the classic Gradient-TD algorithms are not exactly stochastic gradient descent is that they include a product of expectations over the next feature vector in the gradient of the objective function:
\begin{align}
-\frac{1}{2}\nabla\PBE(\vecw) &= \E{(\vecx_t - \gamma \vecx_{t+1})\vecx\tr} \E{\vecx_t\vecx_t\tr}^{-1}\E{\delta_t\vecx_t},\label{eq:GradientOfPBE}\\
&= \Amat\tr\Cmat^{-1} \left(-\Amat\vecw + \vecb\right).\label{eq:GradientOfPBEABC}
\end{align}
The next state's feature vector, $\vecx_{t+1}$, appears in $\Amat$. 
The gradient of $\PBE$ in (\ref{eq:GradientOfPBEABC}) multiplies $\Amat$ with itself and includes a product of expectations of the next state's feature vector. 
To get an unbiased sample of the product, two independent samples are required.
However, during the normal interaction of the agent with the environment, it is only possible to get one sample.
Gradient-TD algorithms get around this issue by learning a second weight vector, $\vecv$, and forming a quasi-stationary estimate of the last two expectations in \eqref{eq:GradientOfPBE}.

Proximal Gradient-TD algorithms solve the double sampling problem by writing the objective function using a saddle-point formulation:
\begin{equation}\label{eq:saddleMSPBE}
\PBE(\vecw) = \min_{\vecw}\max_{\vecv} (\vecb -\Amat \vecw)^\top \vecv - \tfrac{1}{2} \| \vecv \|_\Cmat^2,
\end{equation}
\noindent where $\tfrac{1}{2} \| \vecv \|_\Cmat^2$ is the norm of $\vecv$ weighted by $\Cmat$, or simply $\vecv\tr \Cmat \vecv$. 
Computing the derivative of the objective function (\ref{eq:saddleMSPBE}), with respect to $\vecv$, results in:
\begin{equation} 
\nabla_\vecv{\PBE} = \vecb - \Amat \vecw - \Cmat \vecv, \nonumber
\end{equation}
\noindent and the derivative with respect to $\vecw$ results in:
\begin{equation} 
\nabla_\vecw{\PBE} = -\Amat \tr \vecv. \nonumber
\end{equation}
The gradient of the saddle-point formulated objective does not have the product of the $\Amat$ matrix with itself and avoids the double sampling issue.
This means the algorithm that minimizes the saddle-point objective is a true stochastic gradient descent algorithm and as a result, they can make use of algorithms developed for improving the convergence rate of stochastic gradient descent.
Mahadevan et al. (2014) combined stochastic mirror-prox (Juditsky, Nemirovski, \& Tauvel, 20011) to derive a new version of Gradient-TD family, called Proximal GTD2.
Proximal GTD2\la is described by the following equations:
\begin{align}\label{Proximal GTD2}
\vecv_{t+\frac{1}{2}} \leftarrow& ~\vecv_{t} + \alpha_\vecv \biggl[\delta_{t} \vecz_t - (\vecv_{t}^{\tr}\vecx_t)\vecx_t\biggr],\nonumber\\
\vecw_{t+\frac{1}{2}} \leftarrow& ~\vecw_t +\alpha (\vecv_t^{\tr}\vecx_t)\vecx_t -\alpha \gamma_{t+1} (1 - \lambda_{t+1}) (\vecv_t^{\tr}\vecz_{t} ) \vecx_{t+1},\nonumber\\
\delta_{t+\frac{1}{2}} ~\defeq~& R_{t+1} + \gamma_{t+1} \vecw_{t+\frac{1}{2}}^{\tr}\vecx_{t+1} - \vecw_{t+\frac{1}{2}}^{\tr}\vecx_{t},\nonumber\\
\vecv_{t+1} \leftarrow& ~\vecv_{t} + \alpha_\vecv \biggl[\delta_{t+\frac{1}{2}} \vecz_t - (\vecv_{t+\frac{1}{2}}^{\tr}\vecx_t)\vecx_t\biggr],\nonumber\\
\vecw_{t+1} \leftarrow& ~\vecw_t +\alpha (\vecv_{t+\frac{1}{2}}^{\tr}\vecx_t)\vecx_t -\alpha  \gamma_{t+1} (1 - \lambda_{t+1}) (\vecv_{t+\frac{1}{2}}^{\tr}\vecz_{t} ) \vecx_{t+1}.\nonumber
\end{align}

One of the most recent algorithms proposed for off-policy prediction learning is gradient Temporal Difference learning with Regularized Corrections (TDRC) (Ghiassian et al. 2020).
TDRC(0) differs from GTD(0) in its update of the secondary weight vector.
GTD(0)'s second weight vector minimizes the following objective function:
\begin{align}
J(\vecv) &= \left(\vecv\tr\vecx_t - \delta_t\right)^2,
\intertext{taking the derivative with respect to $\vecv$:}
&= 2 (\vecv\tr\vecx_t - \delta_t) \nabla_\vecv (\vecv\tr\vecx_t - \delta_t),\nonumber \\
&= 2 (\vecv\tr\vecx_t - \delta_t) \vecx_t,\nonumber
\end{align}
which results in the following update for the secondary weight vector for GTD(0):
\begin{align}
\vecv_{t+1} \leftarrow& ~\vecv_t + \alpha_\vecv\biggl[\delta_t\vecx_t - (\vecv_t^{\tr}\vecx_t)\vecx_{t}  \biggr]\nonumber.
\end{align}
TDRC(0) uses the same rationale, with the difference that it uses L2 regularization in its objective:
\begin{align}
J(\vecv) &= \left(\vecv\tr\vecx_t - \delta_t\right)^2 + \beta ||\vecv ||_2^2,\nonumber
\intertext{the derivative of which is:}
&= 2 (\vecv\tr\vecx_t - \delta_t) \vecx_t + \beta ||\vecv||.\nonumber
\end{align}
The original TDRC paper suggests setting setting $\beta=1$, which results in the following update for TDRC's secondary weight vector:
\begin{align}
\vecv_{t+1} \leftarrow& ~\vecv_t + \alpha_\vecv\biggl[\delta_t\vecx_t - (\vecv_t^{\tr}\vecx_t)\vecx_{t}   \biggr] - \alpha_\vecv \vecv_t. \label{eq:TDRC(0)_secondary_weight_update}
\end{align}
The intuition for this update is that we need the correction term in the TD update, but we would like the correction to be small, and we control its magnitude by making the secondary weight vector small.
This is exactly what the regularization term in (\ref{eq:TDRC(0)_secondary_weight_update}) does.
For the case of full bootstrapping, the update for the secondary weight vector is:
\begin{align}
\vecv_{t+1} \leftarrow& ~\vecv_t + \alpha_\vecv\biggl[\delta_t\vecz_t - (\vecv_t^{\tr}\vecx_t)\vecx_{t}  \biggr] - \alpha_\vecv\vecv_t, \nonumber
\end{align}
which similar to the full bootstrapping case, subtracts a multiple the secondary weight vector from the update to keep the magnitude of $\vecv$ small.
The update rule for the primary weight vector is the same as GTD\la.

We close this section by mentioning that there are other objective functions that can be minimized. One alternative objective function is the Mean Squared Bellman Error ($\BE$).
$\BE$ is similar to the $\PBE$ objective but does not have the projection operator.
The so-called \emph{residual gradient} algorithms minimize $\BE$.
There has been some recent work in minimizing the $\BE$ objective using stochastic gradient descent.
These algorithms are true stochastic gradient descent algorithms and in turn provide strong convergence guarantees, however, we do not focus on this class of algorithms in this paper as the $\BE$ is not learnable.
(See Sutton \& Barto, 2018 for a thorough discussion of why minimizing $\BE$ might not be desirable compared to other alternatives).


\subsection{Emphatic-TD Algorithms}
\label{subapp:EmphaticTDAlgorithms}

Emphatic-TD algorithms (Sutton, Mahmood, \& White, 2016) provide an alternative strategy for stable off-policy learning. 
The strategy that Emphatic-TD algorithms use is completely different from the one used by Gradient-TD algorithms. 
Gradient-TD algorithms correct the semi-gradient updates of TD\la so that the updates are in the direction of the gradient and convergence guarantees follow. 
However, Emphatic-TD algorithms use semi-gradient updates.

The main idea of Emphatic-TD algorithms is to emphasize and de-emphasize the update on different time-steps. 
By using emphasis, Emphatic-TD algorithms assure that selective updating in off-policy learning cannot cause divergence. 
This can be explained by going over the $\vecw-2\vecw$ example.
The problem is that the parameter is updated when the agent moves from $\vecw$ to $2\vecw$ but no updates happen when the agent moves from $2\vecw$ to $\vecw$.
If an algorithm assures updates in parameter happen when moving from $2\vecw$ to $\vecw$, the value of $w$ will decrease and divergence can be prevented.
This is exactly what Emphatic-TD algorithms do.
Emphatic-TD algorithms assure that the value of a state (or parameter vector) will be updated if the state is reachable from another state for which we update the parameter vector. 
In the $\vecw-2\vecw$ example, with Emphatic-TD algorithms, if an update happens when the agent moves from $\vecw$ to $2\vecw$, an update is assured when the agent moves from $2\vecw$ to $\vecw$.

The first Emphatic-TD algorithms was Emphatic TD\la, which is sometimes referred to as ETD\la (Sutton, Mahmood, and White (2016)).
This algorithm has one set of learned weights and one step-size parameter.
This algorithm minimizes the emphatic weighted $\PBE$.
The Emphatic TD\la algorithm is described by the following equations:
\begin{align}
F_t \leftarrow& ~\rho_{t-1}\gamma_t F_{t-1} + 1, \label{eq:FollowonTrace} \\
M_{t}  \leftarrow& \lambda + (1 - \lambda)F_t, \label{eq:Emphasis} \\
\vecz_t \leftarrow& ~\rho_t \left(\gamma_t \lambda \vecz_{t-1} + M_t \vecx_{t}\right), \label{eq:ETDTrace} \\ 
\vecw_{t+1} \leftarrow& ~ \vecw_t +\alpha \delta_t \vecz_t \nonumber,
\end{align}
where $F_t$ in (\ref{eq:FollowonTrace}) is the \emph{followon trace} and $M_t$ is called the \emph{emphasis}. As \l approaches 1, $M_t$ approaches $\lambda=1$. 
At the extreme, when $\lambda=1$, we have $M_t=\lambda=1$ in (\ref{eq:Emphasis}), $M_t$ disappears from (\ref{eq:ETDTrace}), and Emphatic TD(1) will reduce to Off-policy TD(1).
As $\lambda\rightarrow 1$, Emphatic TD\la will probably have a behaviour more like TD\la, which means the largest difference between TD\la and Emphatic TD\la should be expected at $\lambda=0$.

Emphatic TD\la is prone to high variance.
In spite of correcting for the difference between the target and behavior policies at each time-step, Emphatic TD\la also corrects for the differences between the policies in the past.
To do this, it uses a product of importance sampling ratios, as shown in (\ref{eq:FollowonTrace}).
The followon trace can become large over time (even unbounded) and the step-size parameter should be reduced further down to avoid divergence.

Emphatic TD($\lambda,~\beta$) was proposed to reduce the variance of Emphatic TD\la. 
As the name suggests, the algorithm has an extra parameter $\beta$ that provides some control over how quickly the magnitude of $F_t$ grows over time. 
All the update rules for Emphatic TD($\lambda,~\beta$) are the same as the ones for Emphatic TD\la, except for the update to the followon trace:
\begin{equation}
F_t \leftarrow ~\rho_{t-1}\beta F_{t-1} + 1, \label{eq:FollowonTraceForLambdaBeta}
\end{equation}
By comparing (\ref{eq:FollowonTrace}) and  (\ref{eq:FollowonTraceForLambdaBeta}) we see that if $\beta$ is set to a value smaller than $\gamma$, $F_t$ will grow at a slower rate than when $\beta=\gamma$.
If $\beta=0$ in (\ref{eq:FollowonTraceForLambdaBeta}), Emphatic TD($\lambda,~\beta$) reduces to Off-policy TD\la, and if $\beta$ is set to $\gamma$, the algorithm reduces to Emphatic TD\la.

The emphasis idea has use-cases other than assuring convergence under off-policy training.
Although originally proposed for off-policy learning, Emphatic TD\la does not reduce to TD\la even if the target and behavior policies are the same.
In fact, Emphatic TD\la is shown to empirically outperform TD\la in some on-policy experiments (Ghiassian, Rafiee, \& Sutton, 2016).
Another possible use-case of the emphasis idea could be in combination with algorithms such as GTD\la.


\subsection{Algorithms for Fast Off-policy Prediction Learning}
\label{subapp:VariableLambdaAlgorithms}

A common concern with off-policy learning is that large importance sampling ratios might cause high variance.\footnote{We would like to note that, to the best of our knowledge, variance issues due to importance sampling ratios have not been concretely demonstrated in the literature. This concern, therefore, is based on intuition and should be considered a hypothesis rather than a known phenomenon.}
Several methods have been proposed that avoid large importance sampling ratios.
Tree Backup\la (Precup, Sutton, \& Singh, 2000), Retrace\la (Munos et al., 2016), and ABQ\ze (Mahmood, Yu, \& Sutton, 2017) all avoid large importance sampling ratios.
Tree Backup\la, and ABQ\ze avoid explicit use of importance sampling ratios in their update rules.
Retrace\la, simply truncates any importance sampling ratio that happens to be larger than one.

Tree Backup, Retrace, and ABQ can all be seen as Off-policy TD\la with \l generalized from a constant to a function of state and action.
This unification was highlighted by Mahmood, Yu, and Sutton, (2017).
The unification makes it simple to explain all three algorithms: all methods are exactly the same as Off-policy TD\la but each method simply uses a different action-dependent trace function $\lambda: \States \times \Actions \rightarrow [0,1]$.
Simply put, each method sets \l differently at each time step.
All three methods mentioned above were originally proposed for control.
Retrace\la was later extended to prediction setting by Espeholt et al. (2018).
Here we provide an easy way to understand all three algorithms and also present the natural state-value variants of all these algorithms.

We begin by providing the generic Off-policy TD algorithm with action-dependent traces.
The key idea is to set $\lambda_t \defeq \lambda(S_{t-1}, A_{t-1})$ such that $\rho_{t-1} \lambda_t$ is well-behaved. The Off-policy TD($\lambda$) algorithm for this generalized trace function can be written:
\begin{align}
\vecz_t & \leftarrow \gamma_t \rho_{t-1} \lambda_t \vecz_{t-1} + \vecx_{t}\label{eq:OffTDTrace}
, \\
\vecw_{t+1} & \leftarrow \vecw_t +\alpha \rho_t \delta_t \vecz_t \nonumber.
\end{align}
Note that the update rules for Off-policy TD\la are different from the ones provided in Sections~\ref{app:PseudoCodes} and \ref{subapp:OffPolicyTD0}.
These new updates explicitly uses $\rho_t$ in the update to $\vecw_{t+1}$ which contrast the earlier Off-policy TD updates which have $\rho_t$ in the trace.
These two forms are actually equivalent, in that the update to $\vecw$ is exactly the same.
We show this equivalence in Appendix~\ref{app_importancesampling}.
We use this other form here, to more clearly highlight the relationship between $\rho_{t-1}$ and $\lambda_t$.
Using the generic update rule of Off-policy TD\la with variable \l, we can now specify different algorithms, by specifying different implementations of the \l function.

Prediction variant of Tree Backup\la is Off-policy TD\la with $\lambda_t = b_{t-1} \lambda$, for some tuneable constant $\lambda \in [0,1]$. 
Replacing $\lambda_t$ with $b_{t-1} \lambda$ in the eligibility trace update in \eqref{eq:OffTDTrace} simplifies as follows:
\begin{align}
\vecz_t & \leftarrow \gamma_t \frac{\pi_{t-1}}{b_{t-1}} b_{t-1}\lambda \vecz_{t-1} + \vecx_{t}, \nonumber \label{eq:TBTrace}\\
&=\gamma_t \pi_{t-1}\lambda\vecz_{t-1}+\vecx_{t}.
\end{align}

A simplified variant of the Vtrace\la algorithm (Espeholt et al., 2018) can be derived with a similar substitution: $\lambda_t = \min\left(\frac{\bar{c}}{\pi_{t-1}},\frac{1}{b_{t-1}}\right) \lambda b_{t-1}$, where $\bar{c} \in \mathbb{R}^+$ and $\lambda \in [0,1]$ are both tuneable constants. The eligibility trace update becomes:
\begin{align}
\vecz_t & = \gamma_t \min \left(\frac{\bar{c}}{\pi_{t-1}},\frac{1}{b_{t-1}}\right) \lambda b_{t-1} \frac{\pi_{t-1}}{b_{t-1}} \vecz_{t-1} + \vecx_{t} \nonumber\\
&= \gamma_t \min \left(\frac{\bar{c}}{\pi_{t-1}},\frac{1}{b_{t-1}}\right) \lambda \pi_{t-1} \vecz_{t-1} + \vecx_{t} \nonumber\\
&= \gamma_t \min \left(\frac{\bar{c}\pi_{t-1}}{\pi_{t-1}},\frac{\pi_{t-1}}{b_{t-1}}\right) \lambda \vecz_{t-1} + \vecx_{t} \nonumber\\
&= \gamma_t \min \left(\bar{c},\rho_{t-1}\right) \lambda \vecz_{t-1} + \vecx_{t}.
\label{eq:V-traceTrace}
\end{align}
The parameter $\bar{c}$ is used to cap importance sampling ratios in the trace.
Note that it is not possible to recover the full Vtrace\la algorithm in this way. The more general Vtrace\la algorithm uses an additional parameter, $\bar{\rho} \in \mathbb{R}^+$ that caps the $\rho_t$ in the update to $\vecw_{t+1}$: $\min(\bar{\rho},\rho_t) \delta_t \vecz_t$. When $\bar{\rho}$ is set to the largest possible importance sampling ratio, it does not affect $\rho_t$ in the update to $\vecw_t$ and so we obtain the equivalence above. For smaller $\bar{\rho}$, however, Vtrace\la is no longer simply an instance of Off-policy TD\la. 
In our experiments, we investigate this simplified variant of Vtrace\la that does not cap $\rho_t$ and set $\bar{c}=1$ as done in the original Retrace algorithm.

ABTD\ze for $\zeta \in [0,1]$ uses $\lambda_t = \nu_{t-1}b_{t-1}$, with the following eligibility trace update:
\begin{align}
\vecz_t & = \gamma_t \frac{\pi_{t-1}}{b_{t-1}} \nu_{t-1} b_{t-1} \vecz_{t-1} + \vecx_{t} \nonumber \\
&=\gamma_t \nu_{t-1}\pi_{t-1}\vecz_{t-1}+\vecx_{t},\label{eq:ABTDTrace}
\end{align}
with the following scalar parameters to define $\nu_t$
\begin{align*}
\nu_t &\defeq \nu(\psi(\zeta), s_t, a_t) \defeq \min \left(\psi(\zeta),\frac{1}{\max(b(a_t|s_t),\pi(a_t|s_t))}\right),\\
\psi(\zeta) &\defeq 2\zeta\psi_0 + \max(0,2\zeta - 1) (\psi_{\text{max}} - 2\psi_0),\\
\psi_{0} &\defeq \frac{1}{\max_{s, a}\max(b(a|s),\pi(a|s))},\\
\psi_{\textnormal{max}} &\defeq \frac{1}{\min_{s, a}\max(b(a|s),\pi(a|s))}
.
\end{align*}

The convergence properties of all three methods are similar to Off-policy TD\la. 
They are not guaranteed to converge under off-policy sampling with weighting $\mu_b$ and function approximation. 
With the addition of gradient corrections similar to GTD\la, all variable-\l algorithms are convergent.
For explicit theoretical results, see Mahmood, Yu \& Sutton (2017) for ABQ with gradient correction and Touati et al. (2018) for convergent versions of Retrace and Tree Backup.

An alternative way to understand variable-\l algorithms is to assume that the term $\rho_{t-1}$ is always present in the eligibility trace update rule, as it should be. 
However, it is not explicit.
In \eqref{eq:TBTrace}, there is no $\rho_{t-1}$, but the term $\pi_{t-1}\lambda$ has the $\rho_{t-1}$ buried inside it:
\begin{align*}
\pi_{t-1}\lambda = \frac{\pi_{t-1}}{b_{t-1}}\lambda_t.
\intertext{We can now remove $\pi_{t-1}$ from both sides of the equation and move $\lambda_t$ to the left hand side:} 
\lambda_t = b_{t-1}\lambda,
\intertext{which is exactly how the effective trace parameter, $\lambda_t$, is set in Tree Backup\la.}
\end{align*}
In the ABTD\ze update rule, \eqref{eq:ABTDTrace} we can similarly write:
\begin{align*}
\nu_{t-1}\pi_{t-1} = \rho_{t-1}\lambda_t.
\intertext{Rearrangement the variables, we come to the effective $\lambda_t$ used by ABTD\ze: }
\lambda_t = \nu_{t-1}b_{t-1}.
\end{align*}
For Vtrace update rule \eqref{eq:V-traceTrace},
\begin{align*}
\min \left(\bar{c},\rho_{t-1}\right) \lambda = \rho_{t-1}\lambda_t,
\end{align*}
\noindent which when rearranged, results in exactly how $\lambda_t$ is set by the Vtrace algorithm:
\begin{align*}
\lambda_t &= \frac{\min\left(\bar{c},\rho_{t-1}\right)}{\rho_{t-1}}\lambda\\
&=\min\left({\bar{c}, \rho_{t-1}}\right) \frac{\lambda}{\rho_{t-1}}\\
&=\min\left(\frac{\bar{c}}{\rho_{t-1}}, 1\right)\lambda\\
&=\min\left(\frac{\bar{c}b_{t-1}}{\pi_{t-1}}, 1\right)\lambda\\
&=\min\left(\frac{\bar{c}}{\pi_{t-1}}, \frac{1}{b_{t-1}}\right)\lambda b_{t-1}.
\end{align*}


\section[Derivations for Tree Backup, ABTD, and Vtrace for Prediction]{Derivations for TB\la, ABTD($\zeta$), and Vtrace\la}
\label{app:DerivationForVariableLambda}

As mentioned before, the three variable-\l algorithms were all originally proposed for control.
The prediction variant of all three algorithms can be derived in a similar way.
To understand the prediction variant of these algorithms, we derive the Off-policy TD\la version of ABQ\ze, which we call ABTD\ze.
Using ABTD\ze, we provide the extensions to Retrace and Tree Backup for prediction.

Consider the generalized \l-return, for a \l based on the state and action---as in ABQ($\zeta$)---or the entire transition (White, 2017).
Let $\lambda_{t+1} = \lambda(S_{t}, A_{t}, S_{t+1})$ be defined based on the transition $(S_{t}, A_{t}, S_{t+1})$, corresponding to how rewards and discounts are defined based on the transition, $R_{t+1} = r(S_{t}, A_{t}, S_{t+1})$ and $\gamma_{t+1} = \gamma(S_{t}, A_{t}, S_{t+1})$.
Then, given a value function $\vhat$, the \l-return $G_t^\lambda$ for generalized $\gamma$ and \l is defined recursively as
\begin{equation}
\Glambda_t \defeq \rho_t \left(R_{t+1} + \gamma_{t+1} \left[ (1-\lambda_{t+1}) \vhat(S_{t+1}) + \lambda_{t+1} \Glambda_{t+1} \right] \right) \nonumber.
\end{equation}
Similar to ABQ($\zeta$) (Mahmood et al., 2017, Equation 7), this \l-return can be written using TD-errors
\begin{equation}
\delta_t \defeq R_{t+1} + \gamma_{t+1} \vhat(S_{t+1}) - \vhat(S_t) \nonumber,
\end{equation}
as
\begin{align*}
\Glambda_t
&= \rho_t \left(R_{t+1} + \gamma_{t+1} \vhat(S_{t+1}) - \gamma_{t+1} \lambda_{t+1} \vhat(S_{t+1}) + \gamma_{t+1} \lambda_{t+1} \Glambda_{t+1} \right)\\
&= \rho_t \left( \delta_t + \vhat(S_{t}) + \gamma_{t+1} \lambda_{t+1} \left[\Glambda_{t+1} - \vhat(S_{t+1})\right] \right)\\
&= \rho_t \delta_t + \rho_t\vhat(S_{t}) + \rho_t \gamma_{t+1} \lambda_{t+1} \left(\rho_{t+1} \delta_{t+1} + \rho_{t+1} \gamma_{t+2} \lambda_{t+2} \left[\Glambda_{t+2} - \vhat(S_{t+2})\right] \right) \\
&= \rho_t \sum_{n=t}^\infty (\rho_{t+1} \lambda_{t+1} \gamma_{t+1})^n \delta_t + \rho_t \vhat(S_{t}),
\end{align*}
where
\begin{equation}
\left(\rho_{t+1} \lambda_{t+1} \gamma_{t+1} \right)^n \defeq \prod_{i=t+1}^n \rho_i \lambda_i \gamma_i
.\nonumber
\end{equation}

This return differs from the return used by ABQ\ze, because it corresponds to the return from a state, rather than the return from a state and action.
In ABQ\ze, the goal is to estimate the action-value for a given state and action.
For ABTD\ze, the goal is to estimate the value for a given state.
For the return from a state $S_t$, we need to correct the distribution over actions $A_t$ with importance sampling ratio $\rho_t$.
For ABQ\ze, the correction with $\rho_t$ is not necessary, and importance sampling corrections only need to be computed for future states and actions, with $\rho_{t+1}$ onward.
For ABTD\ze, therefore, unlike ABQ\ze, not all importance sampling ratios can be avoided.
We can, however, still set \l in a similar way to ABQ\ze to mitigate the variance effects of importance sampling, resulting in the below ABTD\ze algorithm.

The trace function in ABTD\ze is set to ensure $\rho_t \lambda_{t+1}$ is well-behaved.
For some constant $\psi > 0$, let
\begin{align}
\lambda(S_{t}, A_{t}, S_{t+1}) = \nu(\psi, S_t, A_t) b(S_t, A_t)\nonumber\\
\intertext{where}
\nu(\psi, S_t, A_t) \defeq \min \left(\psi, \frac{1}{\max(b(S_t, A_t), \pi(S_t, A_t)}\right) \nonumber
.
\end{align}
In the \l-return, then
\begin{equation*}
\rho_t \lambda_{t+1} =  \frac{\pi(S_t, A_t)}{b(S_t,A_t)} \nu(\psi, S_t, A_t) b(S_t, A_t) = \nu(\psi, S_t, A_t) \pi(S_t, A_t)
.
\end{equation*}
This removes the importance sampling ratios from the eligibility trace.
The resulting ABTD\ze algorithm can be written as the standard Off-policy TD\la algorithm, for a particular setting of \l as explained in \ref{subapp:VariableLambdaAlgorithms}.
The Off-policy TD\la algorithm, with this \l, is called ABTD($\zeta$), with updates
\begin{align*}
\delta_t ~\defeq~& \rho_t \bigl( R_{t+1} + \gamma_{t+1}\vecw_t^{\tr}\vecx_{t+1} - \vecw_t^{\tr}\vecx_{t}\bigr)\nonumber\\
\vecz_t \leftarrow& ~\gamma_t \nu_{t-1} \pi_{t-1} \vecz_{t-1} + \vecx_{t}\textnormal{\quad with } \vecz_{-1} = \bf{0}\\
\vecw_{t+1} \leftarrow& ~ \vecw_t +\alpha \rho_t \delta_t \vecz_t.
\end{align*}

Finally, we can adapt Retrace\la and Tree Backup\la for policy evaluation.
Mahmood, Yu, \& Sutton (2017) showed that Retrace\la can be specified with a particular setting of $\nu_t$ (in their Equation 36).
We can similarly obtain Retrace\la for prediction by setting
\begin{align*}
\nu_{t-1} &= \psi \min \left(\frac{1}{\pi_{t-1}}, \frac{1}{b_{t-1}}\right),
\end{align*}
For Tree Backup\la, the setting for $\nu_t$ is any constant value in $[0,1]$ (see Algorithm 2 of Precup, Sutton \& Singh, (2000)).


\noindent \textbf{An alternative but incorrect extension of ABQ\ze to ABTD\ze}\label{app_abq}
The ABQ\ze algorithm specifies \l to ensure that $\rho_t \lambda_t$ is well-behaved, whereas we specified \l so that $\rho_t \lambda_{t+1}$ is well-behaved.
This difference arises from the fact that for action-values, the immediate reward and next state are not re-weighted with $\rho_t$.
Consequently, the \l-return of a policy from a given state and action is
\begin{equation}
R_{t+1} + \gamma_{t+1} \left[ (1-\lambda_{t+1}) \vhat(S_{t+1}) + \rho_{t+1} \lambda_{t+1} \Glambda_{t+1} \right]
.\nonumber
\end{equation}
To mitigate variance in ABQ\ze when learning action-values, therefore, $\lambda_{t+1}$ should be set to ensure that $\rho_{t+1} \lambda_{t+1}$ is well-behaved.
For ABTD\ze, however, $\lambda_{t+1}$ should be set to mitigate variance from $\rho_t$ rather than from $\rho_{t+1}$.

To see why more explicitly, the central idea of these algorithms is to avoid importance sampling altogether: this choice ensures that the eligibility trace does not include importance sampling ratios.
The eligibility trace $\vecz^a_t$ in TD when learning
action values is
\begin{equation}
\vecz^a_t = \rho_{t} \lambda_{t} \gamma_t \vecz^a_{t-1} + \vecx^a_t,\nonumber
\end{equation}
for state-action features $\vecx^a_t$.
For $\rho_t \lambda_t = \nu_t \pi_t$, this trace reduces to $\vecz^a_t = \nu_{t} \pi_{t} \gamma_t \vecz^a_{t-1} + \vecx^a_t$ (Equation 18, Mahmood et al., 2017).
For ABTD\ze, one could in fact also choose to set $\lambda_{t}$ so that $\rho_t \lambda_t = \nu_t \pi_t$ instead of $\rho_t \lambda_{t+1} = \nu_t \pi_t$. However, this would result in eligibility traces that still contain importance sampling ratios.
The eligibility trace in TD when learning state-values is
\begin{equation}
\vecz_t = \rho_{t-1} \lambda_{t} \gamma_t \vecz_{t-1} + \vecx_t.\nonumber
\end{equation}
Setting $\rho_t \lambda_{t} = \nu_t \pi_t$ would result in the update
$\vecz_t = \rho_{t-1} \nu_t \frac{\pi_t}{\rho_t} \gamma_t \vecz_{t-1} + \vecx_t$, which does not remove important sampling ratios from the eligibility trace.
Rather, the corresponding update for policy evaluation requires $\rho_{t-1} \lambda_{t} = \nu_{t-1} \pi_{t-1}$, giving the ABTD\ze in Section \ref{subapp:VariableLambdaAlgorithms}.


\section{Derivations for Importance Sampling Placement}\label{app_importancesampling}
We first show that the placement of the importance sampling correction term, $\rho$, is equivalent between two conventions found in the literature.
The original work on Off-policy TD($\lambda$) used:

\begin{align*}
\vecw_{t+1} & \leftarrow \vecw_t + \alpha \delta_t \vecz_t \\
\vecz_t^{\rho} & \leftarrow \rho_t\left( \gamma \lambda \vecz_{t-1}^{\rho} + \vecx_{t}  \right)  \textnormal{\quad with } \vecz_{-1}^{\rho} = \bf{0}\\
\delta_t & \defeq R_{t+1} + \gamma \vecw_{t}^{\tr}\vecx_{t+1} - \vecw_{t}^{\tr}\vecx_{t}
\end{align*}
Other works have used a different placement of $\rho$:
\begin{align*}
\vecw_{t+1} & \leftarrow \vecw_t + \alpha \rho_t \delta_t \vecz_t^\prime\\
\vecz_t^\prime & \leftarrow \rho_{t-1} \gamma \lambda \vecz_{t-1}^\prime  + \vecx_{t} \textnormal{\quad with } \vecz_{-1}^\prime = \bf{0}
\end{align*}
Below, we show that given $\vecz_{-1} = \vecz'_{-1} = 0$, the two sets of updates listed above for Off-policy TD\la are equivalent.
We start by showing that $\dt\zt$ is equal to the product $\rho_t\dt\ztp$ at each step, given that $\vecz_{-1}=\vecz_{-1}^\prime=\bf{0}$.
\begin{align}
\rho_t\dt\ztp &= \rho_t\dt \left[ \rho_{t-1} \gamma\lambda \vecz_{t-1}^\prime + \vecx_t \right] \nonumber \\
&=\rho_t\dt \left[ \rho_{t-1} \gamma \lambda ( \rho_{t-2}\gamma \lambda \vecz_{t-2}^\prime + \vecx_{t-1}) + \vecx_t \right] \nonumber\\
&=\rho_t\dt \left[ \left( \gamma \lambda \right)^2 \rho_{t-1} \rho_{t-2} \vecz_{t-2}^\prime + \gamma \lambda \rho_{t-1} \vecx_{t-1} + \vecx_t \right] \nonumber\\
&=\rho_t\dt \left[ \left( \gamma \lambda \right)^2 \rho_{t-1} \rho_{t-2}       \left( \rho_{t-3} \gamma \lambda \vecz_{t-3}^\prime + \vecx_{t-2} \right)      + \gamma \lambda \rho_{t-1} \vecx_{t-1} + \vecx_t \right] \nonumber\\
&=\rho_t\dt \left[ \left( \gamma \lambda \right)^3 \rho_{t-1} \rho_{t-2} \rho_{t-3} \vecz_{t-3}^\prime + \left( \gamma \lambda \right)^2 \rho_{t-1} \rho_{t-2} \vecx_{t-2} + \gamma \lambda \rho_{t-1} \vecx_{t-1} + \vecx_{t} \right] \nonumber\\
&\vdots \nonumber\\
&=\rho_t\dt \left( \sum_{i=0}^t{\left( \gamma \lambda \right)^i \vecx_{t-i} \prod_{k=1}^i{\rho_{t-k}}} \right) + \rho_t\dt \left( \left( \gamma \lambda \right)^{t+1} \prod_{k=1}^{t+1}{\rho_{t-k}} \right)\vecz_{-1}^\prime \nonumber,
\intertext{assuming that $\vecz_{-1}^\prime=\bf{0}$:}
\rho_t\dt\ztp &=\rho_t\dt \left( \sum_{i=0}^t{\left( \gamma \lambda \right)^i \prod_{k=1}^i{\rho_{t-k}}} \right)\label{eq:RhoPrimeZPrimeResults}.
\intertext{On the other hand we have:}
\dt\zt^{\rho} &= \dt \left[ \rho_t \left( \gamma \lambda \vecz_{t-1}^\rho + \vecx_t \right) \right] \nonumber\\
&= \rho_t\dt \left[ \gamma \lambda \rho_{t-1} \left( \gamma \lambda \vecz_{t-2}^\rho+\vecx_{t-1} \right)  + \vecx_{t} \right] \nonumber\\
&= \rho_t\dt \left[ \left( \gamma \lambda \right)^2 \rho_{t-1} \vecz_{t-2}^\rho + \gamma \lambda \rho_{t-1} \vecx_{t-1} + \vecx_{t} \right] \nonumber\\
&= \rho_t\dt \left[ \left( \gamma \lambda \right)^2 \rho_{t-1} \left( \rho_{t-2} \left( \gamma \lambda \vecz_{t-3}^{\rho} + \vecx_{t-2} \right) \right) + \gamma \lambda \rho_{t-1} \vecx_{t-1} + \vecx_{t} \right] \nonumber\\
&= \rho_t\dt \left[ \left( \gamma \lambda \right)^3 \rho_{t-1} \rho_{t-2} \vecz_{t-3}^{\rho} + \left( \gamma \lambda \right)^2 \rho_{t-1} \rho_{t-2} \vecx_{t-2} + \gamma \lambda \rho_{t-1} \vecx_{t-1} + \vecx_{t} \right] \nonumber\\
&\vdots \nonumber\\
&=\rho_t\dt \left( \sum_{i=0}^t{\left( \gamma \lambda \right)^i \vecx_{t-i} \prod_{k=1}^i{\rho_{t-k}}} \right) + \rho_t\dt \left( \left( \gamma \lambda \right)^{t+1} \prod_{k=1}^{t}{\rho_{t-k}} \right)\vecz_{-1}^{\rho} \nonumber
\intertext{which is equal to \eqref{eq:RhoPrimeZPrimeResults} given that $\vecz_{-1}^\rho=\bf{0}$.\nonumber}
\end{align}

Next we show that the third update---which only corrects a part of the TD-error---is unbiased and equal to the two previous sets of updates mentioned above in expectation (but not necessarily at each time step):
\begin{align*}
\vecw_{t+1} & \leftarrow \vecw_t + \alpha \delta_t^{'} \vecz_t^{'}\\
\vecz_t^{'} & \leftarrow \rho_{t-1} \gamma \lambda \vecz_{t-1}^{'}  + \vecx_{t} \textnormal{\quad with } \vecz_{-1}^{'} = \bf{0}\\
\delta_t^{'} & \defeq \rho_t \left( R_{t+1} + \gamma \vecw_{t}^{\tr}\vecx_{t+1}  \right) - \vecw_{t}^{\tr}\vecx_{t}
\end{align*}
These update rules are also valid because they are equal to the previous ones in expectation. That is, $\CEb{\dt\zt}{\H_t}=\CEb{\rho_t\dt\ztp}{\H_t}=\CEb{\dtp\ztpp}{\H_t}$ where $\H_t$ is the history up to time $t$ ($\H_t=\{S_0, A_0, S_1, A_1, \hdots, S_{t-1}, A_{t-1}, S_t\}$). It is immediate from what we showed above that $\CEb{\dt\zt^\rho}{\H_t}=\CEb{\rho_t\dt\ztp}{\H_t}$. Here we show that $\CEb{\rho_t\dt\ztp}{\H_t}=\CEb{\dtp\ztpp}{\H_t}$:

\begin{align}
\CEb{\rho_t\dt\ztp}{\H_t} &= \ztp\CEb{\rho_t\dt}{\H_t} \qquad\qquad\text{(since all of $\ztp$ is part of the history and known)} \nonumber \\
&=\ztp\CEb{\rho_{t} \left( R_{t+1} + \gamma \vecw_{t}^{\tr}\vecx_{t+1} - \vecw_{t}^{\tr}\vecx_{t} \right)}{\H_t} \nonumber\\
&=\ztp\CEb{\rho_{t} \left( R_{t+1} + \gamma \vecw_{t}^{\tr}\vecx_{t+1} \right)}{\H_t} -\CEb{\rho_t \left( \vecw_{t}^{\tr}\vecx_{t}\right)}{\H_t}. \label{eq:deltaPrimezPrime}
\intertext{On the other hand:}
\CEb{\dtp\ztp}{\H_t} &= \ztp\CEb{\dtp}{\H_t} \nonumber \\
&=\ztp\CEb{\rho_{t} \left( R_{t+1} + \gamma \vecw_{t}^{\tr}\vecx_{t+1} \right) - \vecw_{t}^{\tr}\vecx_{t}}{\H_t} \nonumber\\
&=\ztp\CEb{\rho_{t} \left( R_{t+1} + \gamma \vecw_{t}^{\tr}\vecx_{t+1} \right)}{\H_t}-\CEb{\vecw_{t}^{\tr}\vecx_{t}}{\H_t} \label{eq:deltaDoublePrimezPrime}
\intertext{which is equal to \eqref{eq:deltaPrimezPrime} if:}
\CEb{\vecw_{t}^{\tr}\vecx_{t}}{\H_t}&=\CEb{\rho_t \left( \vecw_{t}^{\tr}\vecx_{t}\right)}{\H_t}. \nonumber
\intertext{This is true since $\CEb{\vecw_{t}^{\tr}\vecx_{t}}{\H_t}=\vecw_{t}^{\tr}\vecx_{t}$:} \nonumber
\CEb{\rho_t \left( \vecw_{t}^{\tr}\vecx_{t}\right)}{\H_t}&=\vecw_{t}^{\tr}\vecx_{t} \underbrace{\CEb{\rho_t}{\H_t}}_{=1}=\vecw_{t}^{\tr}\vecx_{t}.
\end{align}


\newpage

\section{Additional Results and Experimental Details}
\label{app:AdditionalResults}

In this section, additional results from the Collision task are provided.
These extra results do not change the conclusions made in the main text; they are merely provided for completeness.


\subsection{Additional Results of Gradient-TD Algorithms}
\label{subapp:AdditionalResultsOfGradientTDAlgorithms}

\begin{figure}[t]
      \centering
      \includegraphics[width=0.9\linewidth]{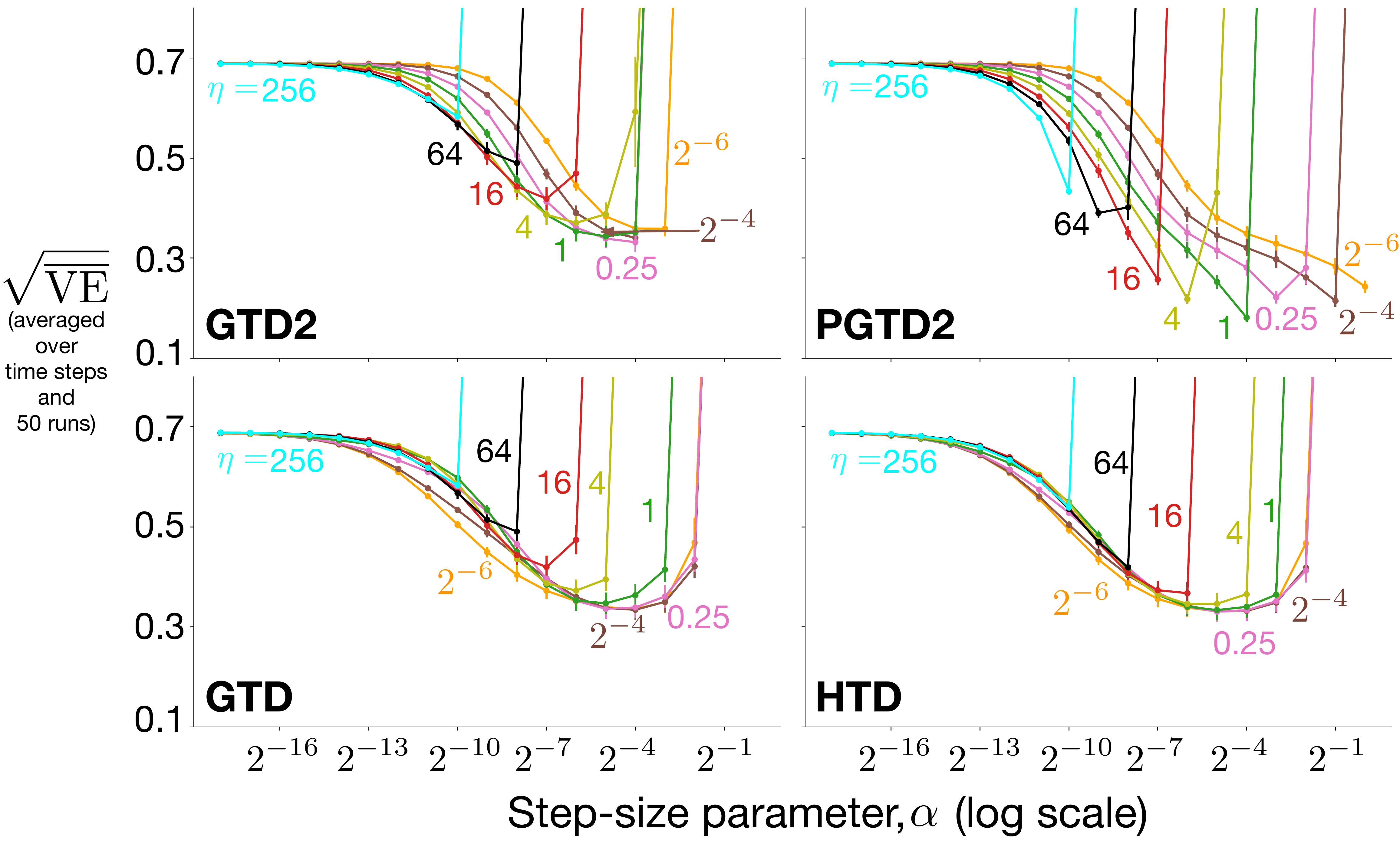}
      \caption{Performance of Gradient-TD algorithms with eight values of $\eta$.}
      \label{fig:7-Collision-Gradients-All-Eta}
\end{figure}
\begin{figure}[b]
      \centering
      \includegraphics[width=0.9\linewidth]{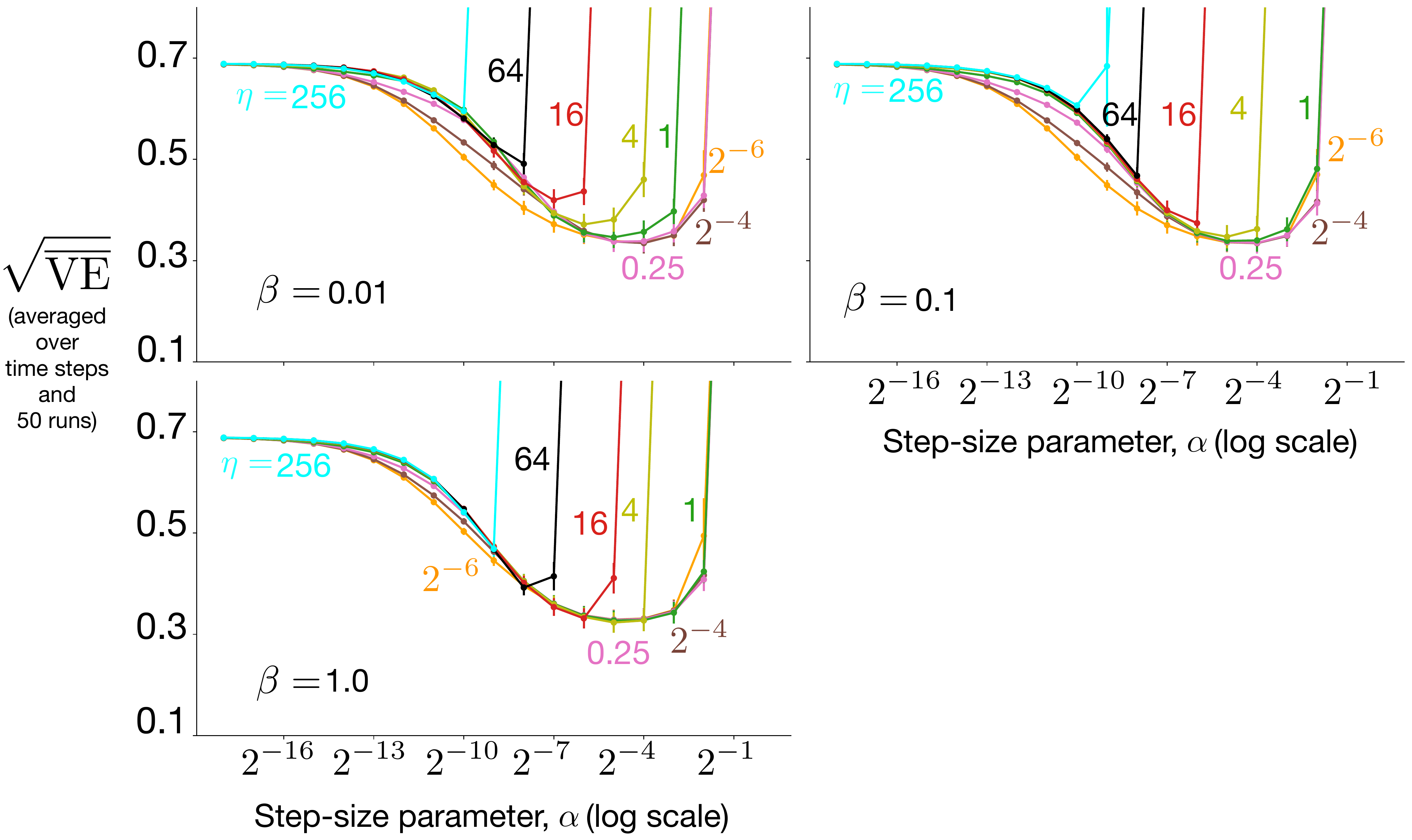}
      \caption{TDRC's performance for various values of $\eta$ and $\beta$.}
      \label{fig:9-Collision-TDRC-Sensitivity-Curves}
\end{figure}

We presented Gradient-TD results with four values of $\eta$ in Section~\ref{sct:AssessmentOfGradietTDAlgorithms}.
Here, we provide the results with eight values of $\eta$ that were applied to the task.
See Figure~\ref{fig:7-Collision-Gradients-All-Eta}.
In general, all algorithms had a low average $\RVE$ with small values of $\eta$.

Additional results for the TDRC\la algorithm are shown in Figure~\ref{fig:9-Collision-TDRC-Sensitivity-Curves}.
In the main text, we restricted TDRC\la results to case with $\eta=1$ and $\beta=1$, where $\beta$ is TDRC's regularization parameter (see Section~\ref{app:PseudoCodes} for TDRC\la update rules).
Sensitivity curves for eight different values of $\eta$ and three values of $\beta$ are shown in Figure~\ref{fig:9-Collision-TDRC-Sensitivity-Curves}.
Each panel of the figure shows the performance for one value of $\beta$.
Each curve in each of the panels shows the performance for one value of $\eta$.
Value of $\beta$ is written inside each panel.
Consistent with the conclusions made in the original paper (Ghiassian et al., 2020), TDRC was fairly robust to the choice of $\beta$.


\subsection{Additional learning curves for various values of the bootstrapping parameter}
\label{subapp:AdditionalLearningCruvesForVariousValuesOfBootstrappingParam}

One of the important attributes of an algorithm is its asymptotic error level.
Here, we show that asymptotic error levels vary and depend on both the algorithm, and the bootstrapping parameter.

\begin{figure}[b]
      \centering
      \includegraphics[width=\linewidth]{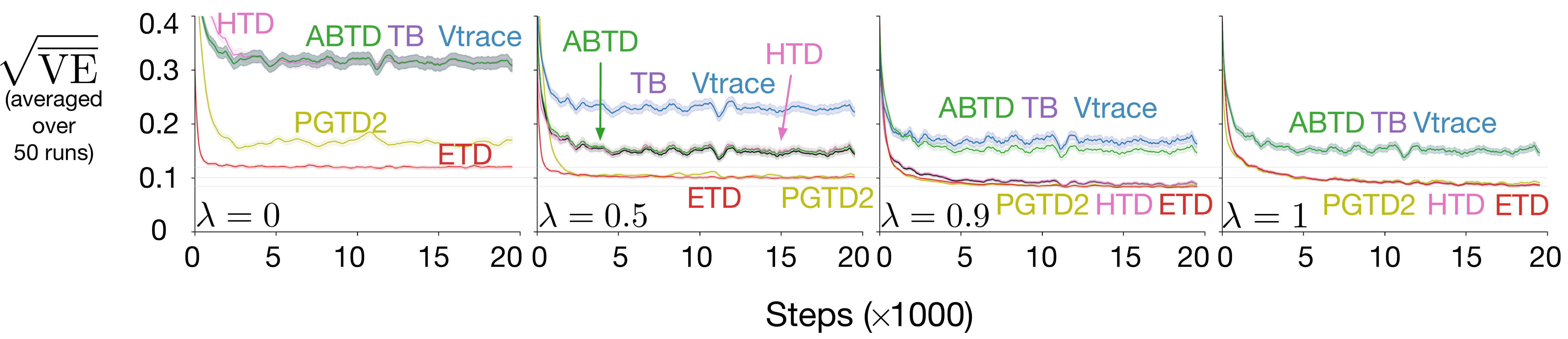}
      \caption{
      Asymptotic error levels of various algorithms.
      }
      \label{fig:10_3-Collision-Learning-Curve-various-lambda}
\end{figure}

Learning curves for algorithms with four values of \l are shown in Figure~\ref{fig:10_3-Collision-Learning-Curve-various-lambda}.
All learning curves in the figure correspond to the algorithm instance that minimizes the area under the learning curve.
The leftmost panel shows the performance of algorithms with full bootstrapping.
All learning curves are averaged over 50 runs and the shaded regions show one standard error. 
Emphatic TD converged to the lowest error level, followed by proximal GTD2.
All other algorithms, including HTD, ABTD, Tree Backup, Vtrace, and the rest that are not shown in the figure converged to a little above 0.3, except Emphatic TD($\lambda, \beta$) whose asymptotic error level depended on the magnitude of $\beta$ and varied between that of Off-policy TD(0) and Emphatic TD(0).

Performance for $\lambda=0.5$ is shown on the second panel of Figure~\ref{fig:10_3-Collision-Learning-Curve-various-lambda}.
By the end of the run, all methods reached a lower asymptotic error level than the full bootstrapping case.
Vtrace and Tree Backup performed similar to each other.
HTD and ABTD performed similar to each other and had a lower asymptotic error level than Tree backup.
Emphatic TD and Proximal GTD2 reached the lowest asymptotic error level than all other algorithms, however, proximal GTD2 learned more slowly than Emphatic TD.

Performance for minimal bootstrapping with $\lambda=0.9$ is shown on the third panel of Figure~\ref{fig:10_3-Collision-Learning-Curve-various-lambda}.
All algorithms performed better than their instance with $\lambda=0.5$, except ABTD.
ABTD(0.9) performed similar to ABTD(0.5).
With minimal bootstrapping, Tree backup, Vtrace, and ABTD all performed similarly.
Proximal GTD2, HTD , and Emphatic TD all converged to the lowest error level of a little below 0.1.

Performance with no bootstrapping is shown on the rightmost panel of Figure~\ref{fig:10_3-Collision-Learning-Curve-various-lambda}.
The only algorithms that performed better than their algorithm instance with $\lambda=0.9$ were Tree backup and Vtrace.
With no bootstrapping, ABTD, Tree backup, and Vtrace performed the same, but worse than the rest of the algorithms.
This is because these algorithms set their effective \l adaptively, which in turns results in asymptotic bias.
Emphatic TD(1) learned more slowly than Emphatic TD(0.9).

\begin{figure}[]
      \centering
      \includegraphics[width=\linewidth]{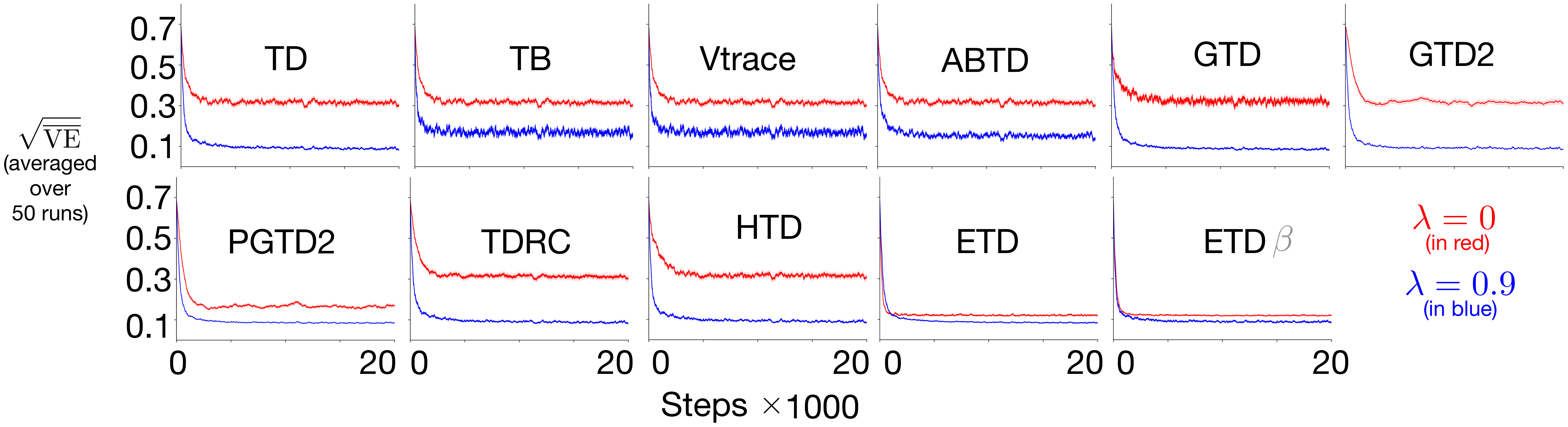}
      \caption{Learning curves for minimal and full bootstrapping.
      }
      \label{fig:4-Collision-Lambda_Learning_Curves}
\end{figure}

To better compare the performance of algorithms with full and minimal bootstrapping, best learning curves (the one with smallest average \MSVE) for all algorithms with full and minimal bootstrapping are plotted in Figure~\ref{fig:4-Collision-Lambda_Learning_Curves}.
Generally, all algorithms performed better with minimal bootstrapping.
They learned faster and converged to a lower error level.
For Emphatic-TD algorithms, little difference in learning speed and final error level was observed with minimal and full bootstrapping.
To avoid the bias induced by maximizing over parameters, we re-ran the algorithm with its best parameters for an extra 50 runs and presented the results for those extra runs.
For more details see Appendix~\ref{subapp:EliminatingParameterMaximizationBias}.


\subsection{A comparison of algorithms with full bootstrapping}
\label{subapp:AComparisonOfAlgorithmsWithFullBootstrapping}

\begin{figure}[]
      \centering
      \includegraphics[width=0.9\linewidth]{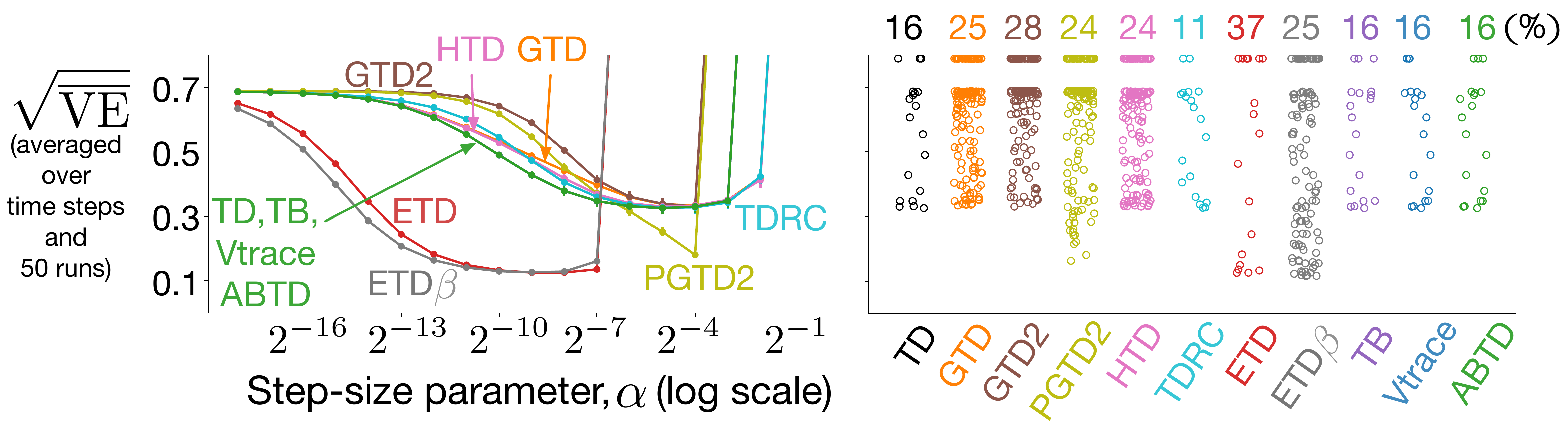}
      \caption{
      Results with full bootstrapping.
      }
      \label{fig:1-Collision-all-algs}
\end{figure}

Here, we make a high level comparison of all algorithms with full bootstrapping.
Sensitivity to step size is shown on the left panel of Figure~\ref{fig:1-Collision-all-algs}.
With full bootstrapping, Emphatic-TD algorithms had the lowest average \MSVE with their appropriate $\alpha$.
This lower error is a result of learning faster and converging to a lower final error level than other algorithms (see Figure~\ref{fig:0-Collision-sample_learning_curves}).
All other algorithms had an error of a little above 0.3, except for proximal GTD2 that reached to a lower error level than others.
TDRC and HTD had the widest sensitivity curves at their bottom, meaning that they were less sensitive to $\alpha$ and in turn were easier to use than other Gradient-TD algorithms.
GTD2 and proximal GTD2 had a higher error than all other algorithms when the step size was around $2^{-10}$.

The difference observed between the two families of Emphatic- and Gradient-TD is statistically significant because it is more than twice the standard error in both means.
Similarly, the difference between proximal GTD2 and other Gradient-TD algorithms is statistically significant.

\emph{Waterfall plots} provide another representation of the results.
Waterfall plots are a great way of presenting \textbf{all} the data points in a condensed manner in one plot.
All the results for the case of full bootstrapping are shown on the right panel of Figure~\ref{fig:1-Collision-all-algs}.
We see that some algorithms have more data points on the waterfall plot than others.
These are the algorithms that have more than one tuned parameter and as a result, more instances of such algorithms were applied to the task.
The numbers at the top of the waterfall plot shows the percentage of parameter settings that learned unstably and had an average \MSVE higher than their \MSVE at $t=0$.
For better visibility, these are shown as having an error of 0.8.


\subsection{Eliminating Bias Incurred by Maximizing Over Parameters}
\label{subapp:EliminatingParameterMaximizationBias}

Wherever we maximized over parameters to find the best parameter setting, we followed the following procedure to eliminate the bias incurred by maximizing over parameters: The parameters that resulted in the lowest error were selected.
With those selected parameters, the specific algorithm instance was applied to the task for an additional 50 times (extra to the original 50 runs that resulted in the original error value) and the results for these extra 50 runs were presented.

\begin{figure}[]
      \centering
      \includegraphics[width=0.8\linewidth]{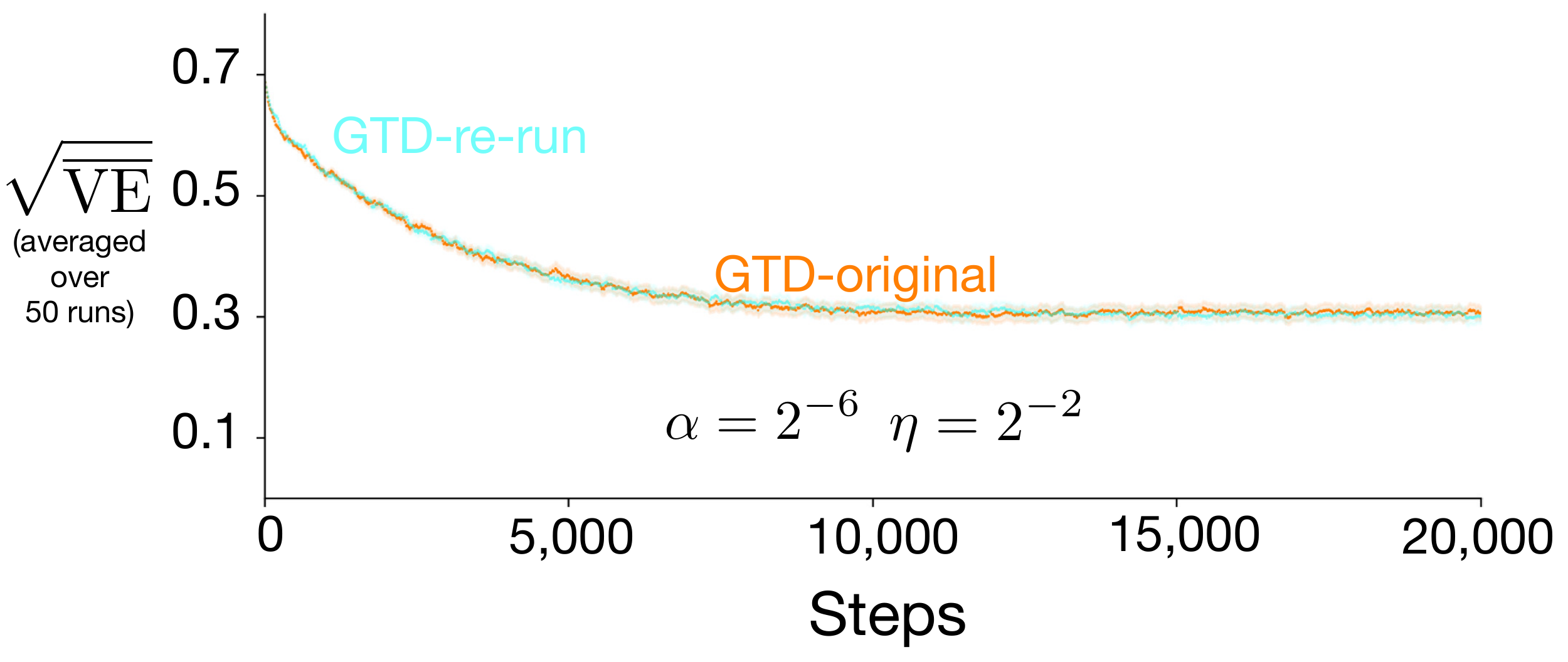}
      \caption{Average error over 50 original and 50 additional runs for GTD(0).}
      \label{fig:8-Collision-Learning-curves-for-original-and-rerun}
\end{figure}

Two learning curves are shown in Figure~\ref{fig:8-Collision-Learning-curves-for-original-and-rerun}.
Both learning curves show the average error over 50 runs for the GTD(0) algorithm with $\alpha=2^{-6}$, and $\eta=2^{-2}$.
Each learning curve is the average error over a different set of 50 runs.
The learning curve in orange is the average error over the original 50 runs.
The results from these runs were used to find the parameter setting that minimized the average error of the algorithm towards the end of the run (last 5\% of the time steps in this case).
After these parameters were found, the algorithm was applied to the task for another 50 runs.
As seen in the figure, in this case, re-running with the best parameters did not result in error levels that were statistically significantly different from the original results.

\subsection{Specification of Dependencies and More Details}
\label{subapp:SpecificationOfDependencies}

Experiments were conducted using a supercomputer with 2,024 nodes. Each node had 2 sockets with 20 Intel Skylake cores (2.4 GHz, AVX512), for a total of 40 CPUs per node and an overall total of 80960 cores.
Each node had 202 GB of RAM.
The operating system used was Linux CentOS 7.

Python 3.6 was used to run the code, with numpy version 1.19.0.
To plot the data, matplotlib version 3.2.2 was used.

For more details, see the code at: \href{https://github.com/sinaghiassian/OffpolicyAlgorithms}{https://github.com/sinaghiassian/OffpolicyAlgorithms}.

\end{document}